\pdfoutput=1

\documentclass[11pt]{article}

\usepackage[final]{acl}

\usepackage{times}
\usepackage{latexsym}

\usepackage{amsmath}
\usepackage{booktabs}
\usepackage{graphicx}
\usepackage{todonotes}
\usepackage{tabularx}
\usepackage{subcaption}
\usepackage{caption}
\usepackage{array}
\usepackage{booktabs}
\usepackage{color, colortbl}
\usepackage{tcolorbox}
\usepackage{threeparttable}
\usepackage[inline]{enumitem}
\newlist{inparaenum}{enumerate*}{1}
\setlist[inparaenum,1]{label=(\arabic*), itemjoin={{ }}, itemjoin*={{ }}}

\usepackage[capitalise]{cleveref}

\usepackage[T1]{fontenc}

\usepackage[utf8]{inputenc}

\usepackage{microtype}

\usepackage{inconsolata}

\usepackage{graphicx}
\usepackage{xspace}

\newcommand{\dataset}{\textsc{PosterSum}\xspace}
\newcommand{\method}{\textsc{Segment \& Summarize}\xspace}

\title{\textsc{PosterSum}: A Multimodal Benchmark for \\ Scientific Poster Summarization} 

\author{Rohit Saxena \qquad Pasquale Minervini \qquad Frank Keller \\
 Institute for Language, Cognition and Computation\\
 School of Informatics, University of Edinburgh \\
 10 Crichton Street, Edinburgh EH8 9AB \\
  \texttt{rohit.saxena@ed.ac.uk} \quad  \texttt{p.minervini@ed.ac.uk} \quad \texttt{keller@inf.ed.ac.uk}}

\begin{document}
\maketitle
\begin{abstract}
Generating accurate and concise textual summaries from multimodal documents is challenging, especially when dealing with visually complex content like scientific posters.
We introduce \textsc{PosterSum}\footnote{The dataset is available at \href{https://huggingface.co/datasets/rohitsaxena/PosterSum}{rohitsaxena/PosterSum}.}\footnote{The code is available \href{https://github.com/saxenarohit/postersum}{at this link}.}, a novel benchmark to advance the development of vision-language models that can understand and summarize scientific posters into research paper abstracts.
Our dataset contains 16{,}305 conference posters paired with their corresponding abstracts as summaries.
Each poster is provided in image format and presents diverse visual understanding challenges, such as complex layouts, dense text regions, tables, and figures.
We benchmark state-of-the-art Multimodal Large Language Models (MLLMs) on \textsc{PosterSum} and demonstrate that they struggle to accurately interpret and summarize scientific posters.
We propose \method, a hierarchical method that outperforms current MLLMs on automated metrics, achieving a 3.14\% gain in ROUGE-L.
This will serve as a starting point for future research on poster summarization.
\end{abstract}
\begin{figure*}[tb]
  \centering
  \vspace{-2ex}
  \includegraphics[width={\textwidth}]{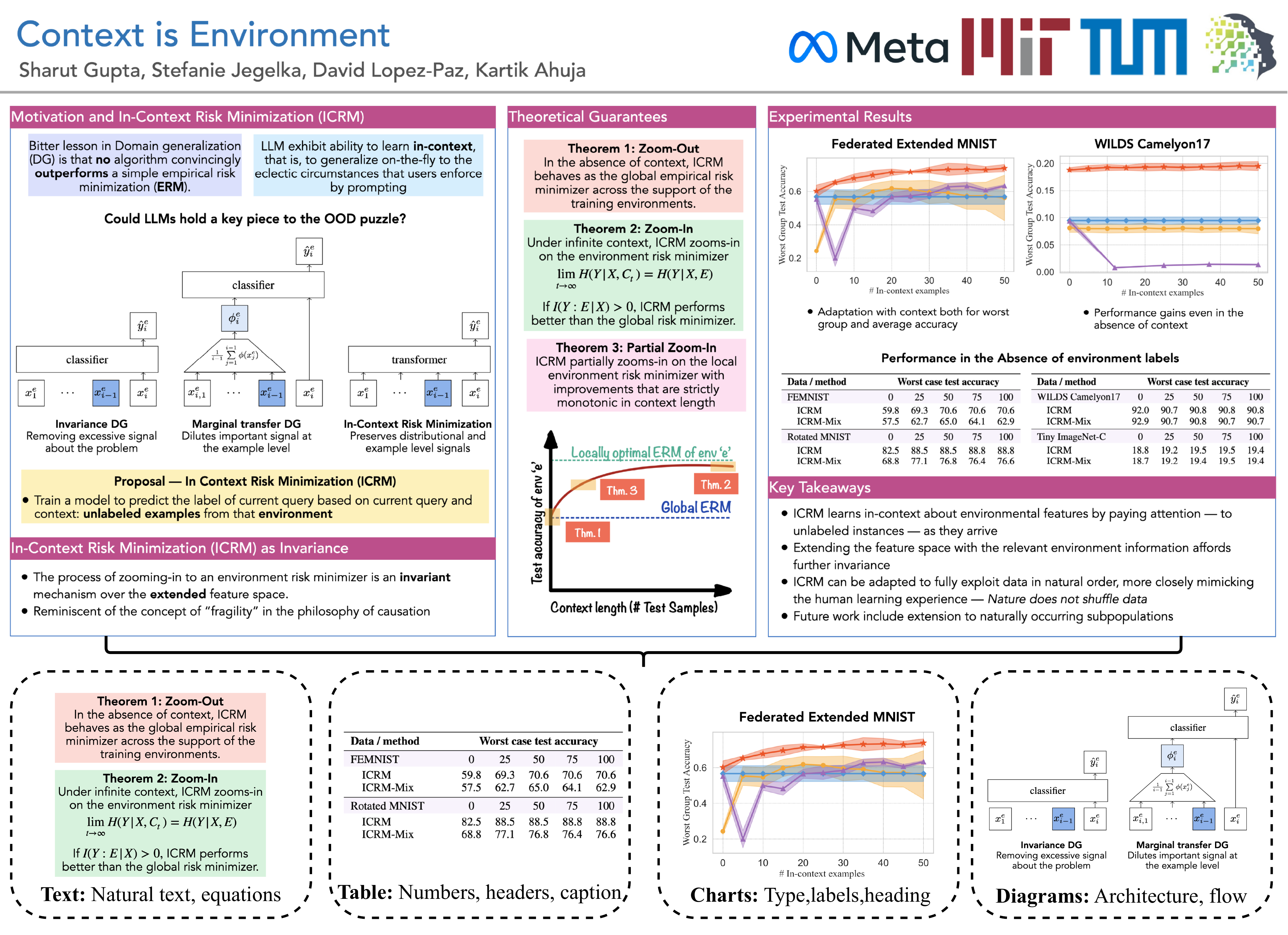}
  \caption{An example of scientific poster from the \textsc{PosterSum} dataset. The poster, describing the work in \citet{GuptaJLA24}, contains visual elements such as structured tables with numerical results, charts, diagrams, and textual sections, demonstrating the multimodal complexity present in the dataset.}
  \label{fig:poster_example}

\end{figure*}
\section{Introduction}
Scientific posters play a critical role in academic communication, offering a visually rich medium that combines text, images, charts, and other graphical elements to present research findings. Summarizing these visually complex posters into concise and accurate textual abstracts presents a unique challenge, requiring models to integrate multimodal information effectively.  

Multimodal Large Language Models ~\citep[MLLMs;][]{openai2024gpt4technicalreport, llama3herd} have demonstrated remarkable capabilities in vision-and-language tasks, including image captioning~\citep{fu2023_MME,Koh_Grounding,Yu_2024_CVPR_CapsFusion,garg-etal-2024-imageinwords} and visual question answering~\citep{MMBench,Yue_2024_CVPR}.
While these models exhibit strong generalization across various domains, their performance often declines when applied to scientific text~\citep{li-etal-2024-multimodal-arxiv,lu2024mathvista,pramanick2024spiqa}.
Additionally, the complexity of poster layouts, the use of technical terminology, and the intricate interplay between text, tables, and figures make summarizing scientific posters a particularly challenging task, which has remained under-explored due to the lack of specialized datasets.

To address this gap, we introduce \textsc{PosterSum}, a novel multimodal benchmark for summarizing scientific posters into research paper abstracts.
Our dataset consists of 16,305 scientific posters and corresponding abstracts as summaries collected from the main Machine Learning conferences, namely ICLR, ICML, and NeurIPS.
%
These posters cover a broad range of scientific disciplines and present unique challenges, including complex layouts and intricate combinations of text, tables, and figures as shown in \cref{fig:poster_example}.
Information is often distributed across the poster, requiring careful navigation and integration of diverse elements to identify and summarize the key points effectively.
%

We benchmark state-of-the-art MLLMs on \textsc{PosterSum} and demonstrate that, despite their impressive performance on a range of other multimodal tasks, these models face significant limitations when summarizing scientific posters.
For instance, the best-performing closed-source model in our experiments, GPT-4o~\citep{openai2024gpt4technicalreport}, achieves a ROUGE-L score of 22.30, underscoring the difficulty of this task specifically with the posters with figures and tables.
To address this challenge, we propose \method, a hierarchical approach inspired by the divide-and-conquer principle~\citep{divide_and_conquer_cvpr}.
The method involves three key steps: (1) Segmentation: we segment each poster into coherent regions; (2) Localized Summarization: a multimodal large language model extracts and interprets the text within each segment, generating localized summaries for each region; and (3) Global Summarization: these localized summaries are combined using text-based large language model to produce a cohesive abstract that spans the entire poster. Notably, this approach does not require additional training or fine-tuning.
Local summaries allow the model to focus on fine details within that specific area, which are useful for tables and figures.
Also, it aligns with the inherent structure of the poster, which has sections with a specific focus.
This approach achieves a ROUGE-L score of 24.18, outperforming both closed-source and open-source models, setting a new benchmark for scientific poster summarization.

The proposed dataset and baselines will enable future research in multimodal scientific poster understanding.
Our contributions can be summarized as follows:
\begin{itemize}[left=0pt,nosep]
\item We introduce \dataset, a large-scale multimodal dataset of 16,305 scientific posters paired with their abstracts, tailored for research poster summarization.
\item We benchmark state-of-the-art MLLMs on \textsc{PosterSum}, showing their limitations in processing and summarizing scientific posters.
\item We propose \method, a hierarchical approach that segments each poster into coherent regions, extracts the textual content from those regions and then composes a final summary; we also demonstrate \dataset's utility for fine-tuning MLLMs, showing promising improvements over zero-shot results.
\end{itemize}

\section{Related Work}

\paragraph{Multimodal Large Language Models.}
%
After the emergence of LLMs, recent work~\citep{DBLP:conf/nips/LiuLWL23a, DBLP:journals/corr/abs-2311-03079, DBLP:conf/nips/AlayracDLMBHLMM22} investigated their use in processing multimodal inputs, giving rise to Multimodal Large Language Models (MLLMs).
The core idea in this line of research is to align visual and textual features by using shared representations.
This framework typically involves using a pre-trained visual encoder to extract visual features, a projection layer to map visual representations into corresponding text representations, and a pre-trained LLM to generate textual responses, allowing the model to condition the output on visual and textual inputs.
MLLM architectures such as LLaVA~\cite{DBLP:conf/nips/LiuLWL23a} and MiniCPM~\cite{yao2024minicpmvgpt4vlevelmllm} demonstrated impressive zero-shot generalization across diverse visual and language tasks.
However, most existing MLLMs focus on general domain tasks and relatively simple visual inputs; the challenge of understanding complex and information-dense visual documents like scientific posters remains under-explored.
\paragraph{Summarization in Scientific Domains.}
\emph{Scientific summarization} consists of generating concise summaries for scientific content~\citep{ScisummNet,DBLP:conf/emnlp/CacholaLCW20,DBLP:conf/emnlp/JuLKJDP21,DBLP:conf/naacl/SotudehG22}.
Several scientific summarization benchmarks have been proposed, 
designed to process modalities such as videos~\cite{DBLP:conf/acl/LevSHJK19,m3av}, slide decks~\cite{DBLP:conf/aaai/TanakaNNHSS23}, surveys~\cite{DBLP:conf/acl/LiuLYZJLH24}, 
and research papers~\cite{DBLP:conf/naacl/TakeshitaGR0P24,DBLP:conf/coling/PuWLD24}. 
However, scientific poster summarization remains unexplored despite the widespread use of posters in academic communication.

\paragraph{Document Layout Analysis and Segmentation.}
Understanding document layouts plays a significant role in processing complex visual documents like scientific posters.
Recent work in document layout analysis~\cite{ERNIE,DBLP:conf/acl/0005RSMBKPNL24,DBLP:conf/cvpr/LuoSZZYY24,DocFormerv2} aims at identifying and classifying different regions within a document considering spatial relationships and content type.
Previous work has also focused on understanding individual elements in documents, such as charts~\citep{masry-etal-2022-chartqa} and tables~\citep{zheng-etal-2024-multimodal}.
However, most existing approaches are designed for either standard documents or individual elements like charts and tables and do not capture the complex layouts and the rich multimodal structure of scientific posters, which typically consist of text, charts, equations, and tables.

\begin{figure}[t]
  \centering
  \vspace{-2ex}
\includegraphics[width=\columnwidth]{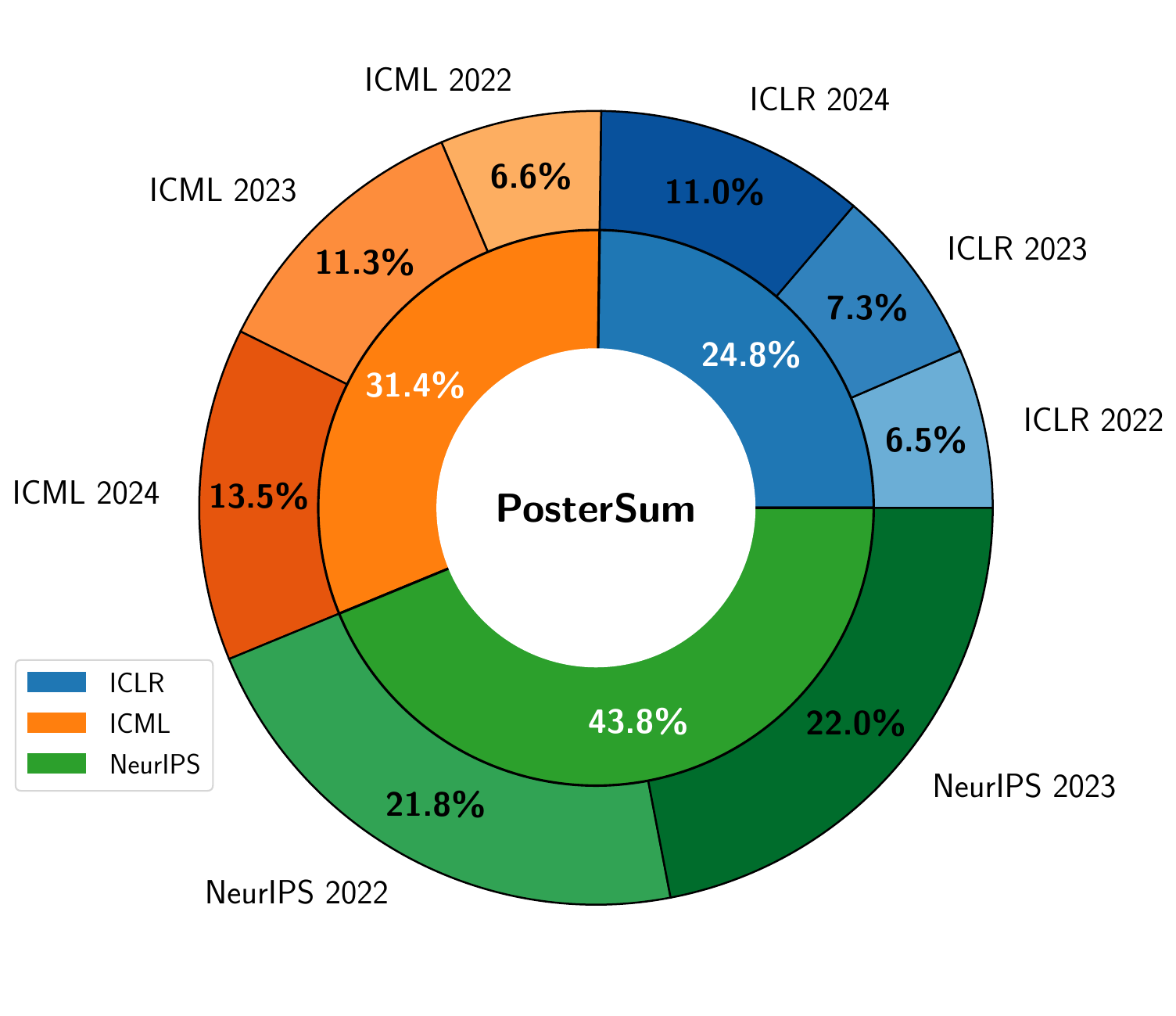}
\vspace{-2ex}
  \caption{Distribution of the \textsc{PosterSum} dataset.}
  \label{fig:year_dist}
\end{figure}
\section{The \dataset Dataset}
We introduce \textsc{PosterSum}, a novel dataset and benchmark for multimodal abstractive summarization of scientific posters. The dataset consists of 16,305 pairs of academic posters as images (PNG format) and their corresponding research paper abstracts. These posters were collected from major machine learning and artificial intelligence conferences, which accept papers from various subfields of machine learning,  including computer vision, natural language processing, optimization, and computational biology. 

\textsc{PosterSum} captures the diverse and heterogeneous nature of academic posters, which are commonly used at conferences to present research findings. These posters vary in layout, content, and visual complexity--some are text-heavy, while others emphasize visual elements such as charts, graphs, and figures, as shown in \cref{fig:poster_example}. This variability presents a significant challenge for MLLMs, requiring them to interpret and summarize multimodal information effectively.

Each poster in the dataset is paired with its corresponding abstract, which serves as the ground-truth summary. The abstract highlights the key contributions and findings of the research, making it an ideal summary for the poster. Unlike image captioning, poster summarization requires a deeper understanding of multiple elements in the poster to generate a comprehensive and meaningful abstract-based summary.
\subsection{Dataset Creation}
The \textsc{PosterSum} dataset was collected from the websites of top-tier machine learning and artificial intelligence conferences: \href{https://iclr.cc/}{ICLR}, \href{https://icml.cc/}{ICML}, and \href{https://neurips.cc/}{NeurIPS}. We selected these conferences based on the availability of research posters. We first collected research paper links and paper identifiers from the conference websites. We filtered out any entries where the poster of the paper was not available, ensuring that only papers with accessible posters were included in the dataset. We exclusively collected posters from the years 2022 to 2024, as shown in \cref{fig:year_dist}. Additionally, we manually reviewed the dataset to remove any posters with placeholder images. We assume that the research reported in the posters is of a high standard, and the posters are of high quality, as the corresponding papers appeared at top machine learning conferences.

To build a robust summarization dataset, it was essential to pair each poster with a human-written summary. We collected the research paper abstracts from the corresponding paper pages using the paper identifiers. These abstracts serve as the summaries for the posters, as they highlight the core findings and contributions of the research. For papers where the abstract was missing from the webpage, we manually extracted the abstract from the research paper’s PDF to ensure completeness.

\begin{figure}[t!]
  \centering
  \vspace{-2ex}
\includegraphics[width=0.9\columnwidth]{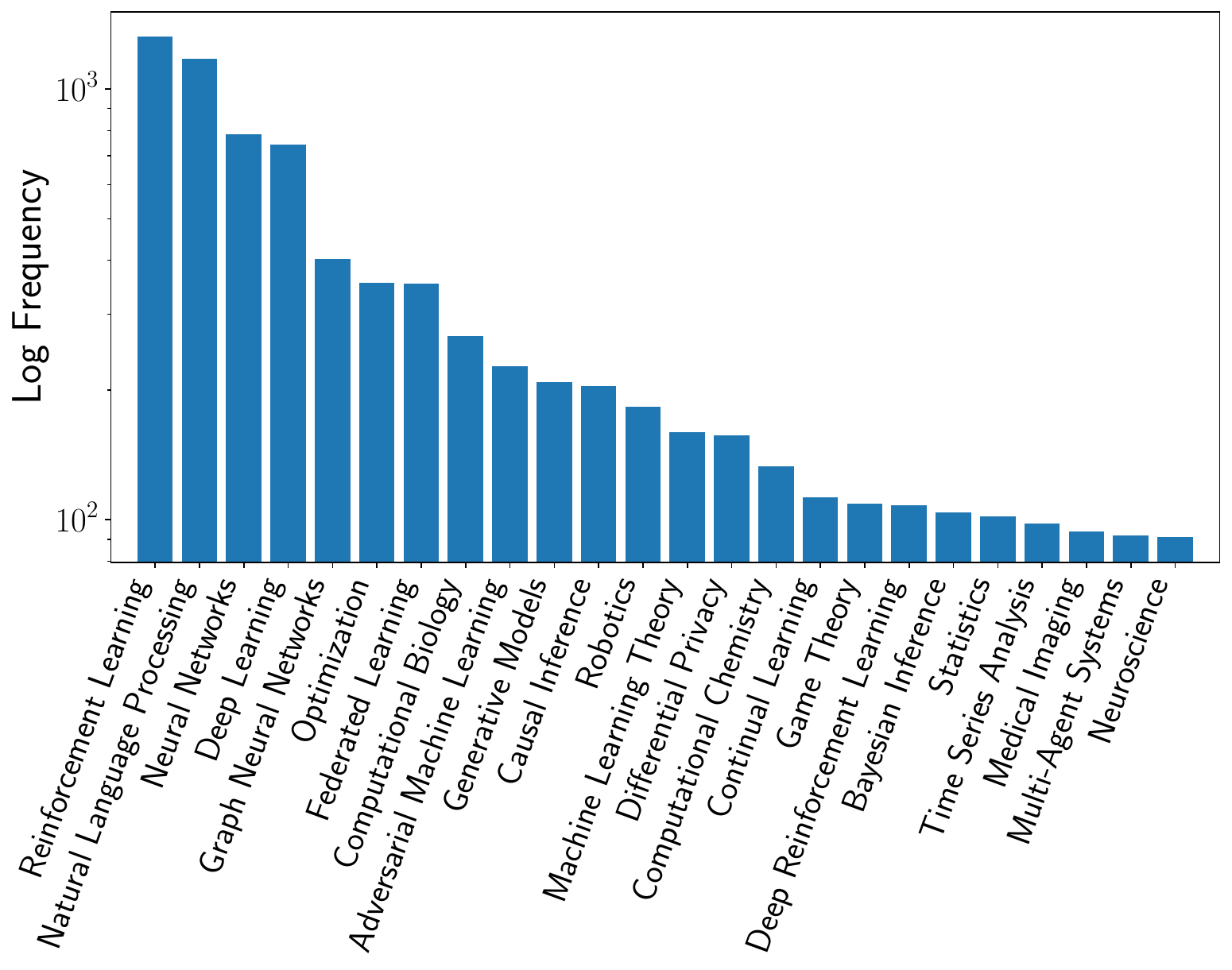}
\vspace{-2ex}
  \caption{Distribution of top 25 topics for the posters in the dataset.}
  \label{fig:topic_dist}
\end{figure}
\subsection{Dataset Statistics and Analysis}
\begin{table}[t]
\centering
\scalebox{0.9}{%
\begin{tabular}{@{}lr@{}}
\toprule
\textbf{\textsc{PosterSum} Statistics} \\
\midrule
Total number of posters-summary & 16,305         \\
Total number of unique categories & 137         \\
Mean token length of the summary  & 224      \\
Mean summary sentences  & 7.21\\ 
Train/Val/Test size& 10305/3000/3000 \\
Mean CLIP score & 29.08\\
Year range & 2022--2024\\
\bottomrule
\end{tabular}}
\caption{Statistics of the \textsc{PosterSum} dataset.}
\label{tab:stats}
\end{table}
\vspace{-1ex}
\begin{table}[t!]
\centering
\scalebox{0.9}{%
\begin{tabular}{@{}cccc@{}}
\toprule
\multicolumn{4}{c}{\bf{\% Novel n-grams in Summary}} \\
\bf{1-grams} & \bf{2-grams} & \bf{3-grams} & \bf{4-grams} \\
\midrule
54.54 & 81.13 & 88.67 & 91.41 \\
\bottomrule
\end{tabular}}
\caption{Statistics for percentage of novel n-grams in the \textsc{PosterSum} summaries.}
\label{tab:ngram}
\end{table}
This process resulted in the 16,305 poster-summary pairs, providing a comprehensive multimodal resource for evaluating abstractive summarization of academic research posters.

\cref{tab:stats} provides an overview of key statistics for the dataset. The average length of the poster summaries is 224 word-piece tokens, with an average of seven sentences per summary. The poster images are of high-resolution, with a mean size of $3547 \times 2454$. We randomly split the dataset into training, validation, and test sets using a 10305/3000/3000 split, which can be utilized for training and fine-tuning models.

To better understand the diversity within the dataset, we categorized each poster into topics. Since topics were not available on the conference websites, we employed the GPT-4o vision model to generate topic labels by prompting the model in a zero-shot setting using the images of the posters. As a result, we identified 137 distinct topics within machine learning and artificial intelligence for the posters, spanning areas such as reinforcement learning, natural language processing (NLP), computational biology, and healthcare applications. \cref{fig:topic_dist} illustrates the distribution of the top 25 topics by frequency.

To assess the abstractiveness of the poster summaries, we report the percentage of novel n-grams in the summaries compared to the Optical Character Recognition (OCR) extracted text from the posters. We used MMOCR~\citep{MMOCR} to extract the text.
While most posters do not explicitly include abstracts, we found that approximately 8\% of the total posters may contain an abstract in poster, based on the occurrence of the word "abstract" in the OCR text.
As shown in \cref{tab:ngram}, a significant portion of the summaries contains novel content, particularly in the 3-gram and 4-gram categories. 
This demonstrates that the summaries are not simple restatements of poster text but instead provide a more comprehensive abstraction.

We also find a mean CLIP score~\cite{hessel-etal-2021-clipscore} of 29.08 when we evaluate the alignment between the images of the posters and their summaries.
This score was computed at the sentence level and averaged across the dataset. The relatively low CLIP score highlights the challenge that \textsc{PosterSum} poses for existing MLLMs. Unlike image-captioning tasks, where captions directly describe visual features, academic posters are composed of diverse and complex visual elements, such as charts, graphs, equations, and dense textual explanations. This complexity makes it more difficult for models to capture the semantic relationships between these elements and the corresponding abstract summaries.
\section{Multimodal Poster Summarization}
\subsection{Task Formulation}
Given a scientific poster $I$ in image format as input, the objective is to generate a textual summary $\hat{Y}=\{\hat{y}_1,\hat{y}_2,\dots,\hat{y}_m\}$ that encapsulates the key points and essential content of the poster.
Formally, a model $M_\theta$, parameterized by $\theta$, takes the poster $I$ as input, optionally accompanied by a prompt $P$, and generates a summary $\hat{Y}$. The key challenge in this task is that model $M_\theta$ must effectively abstract from the diverse visual and textual elements present in the poster, including text, charts, diagrams, and equations, to produce a coherent and informative summary.

\subsection{Baselines}
We evaluate various multimodal models, both open-source and closed-source, to assess their performance on the abstractive summarization task for scientific posters.
As the posters include textual elements, we also evaluate OCR-based methods as baselines.
For MLLMs, evaluation is conducted in a zero-shot and Chain-of-Thought (CoT) setting to assess the capability of models to generate accurate summaries.
Additionally, we explore parameter-efficient fine-tuning techniques on selected open-source models.
Below are the categories of models used in our experiments.
\paragraph{Optical Character Recognition (OCR).}
For OCR-based baselines, we used two OCR methods (MMOCR~\citep{MMOCR} and Pytesseract\footnote{\url{https://github.com/h/pytesseract}}) to extract text from the poster images and concatenated the results to generate a summary. Additionally, we combined the best OCR output with a text-based large language model (LLM). In this approach, we first extract text from the posters and then use the Llama-3.1-8B-Instruct~\citep{llama3herd} model for summarization. This allows us to evaluate the performance of text-only LLMs when provided with OCR-extracted text.

\paragraph{Closed-source MLLMs.}
We evaluated GPT-4o~\citep{openai2024gpt4technicalreport}, Claude 3.5 Sonnet~\citep{anthropic_claude3_5_sonnet}, and Gemini 2.0~\citep{geminiteam2024} as closed-source MLLMs.
All the models were prompted with the image of the poster in a zero-shot setting to generate abstractive summaries based on the input. The prompt template can be found in \cref{sec:templates}.
\paragraph{Open-source MLLMs.}
As open-source/open-weights models, we evaluated Llama-3.2-11B-Vision-Instruct~\citep{llamaV}, Qwen2-VL-7B-Instruct~\citep{qwen2technicalreport}, LLaVA-NeXT~\citep{liu2024llavanext,liu2023improvedllava}, mPLUG-DocOwl2~\citep{hu2024mplugdocowl2highresolutioncompressingocrfree}, and MiniCPM-Llama3-V-2.5~\citep{yao2024minicpmvgpt4vlevelmllm}. Each model was evaluated in both zero-shot and CoT settings. The CoT prompt was used to steer the models to extract relevant information, such as the title, research problem, methods, results, and conclusion, from the poster. We report the full prompt template in \cref{sec:templates}.

\paragraph{Fine-tuned Models (LoRA).} We also evaluated the fine-tuned Llama-3.2-11B-Vision-Instruct and LLaVA-NeXT models. We used parameter-efficient fine-tuning using the Low-rank Adaptation ~\citep[LoRA;][]{LoRA} method to fine-tune both of these models using the training and validation set of the \dataset dataset. 
\begin{table*}[tb]
\centering
\scalebox{0.9}{
\begin{tabular}{@{}lrrrrrrrrr@{}} 
\toprule            
 & \textbf{R-1} & \textbf{R-2} & \textbf{R-L} & \textbf{RLSum} & \textbf{SBLEU} & \textbf{Met} & \textbf{BS\textsubscript{p}} & \textbf{BS\textsubscript{r}} & \textbf{BS\textsubscript{f1}} \\ 
\midrule
\textbf{Closed-Source Models} & \multicolumn{7}{@{}l}{} \\
\midrule
Gemini & 39.89 & 12.38 & 20.89 & 36.21 & 6.57 & 22.34 & 59.46 & 59.6 & 59.53\\
Claude-3.5 Sonnet & 43.45 & 11.42 & 19.51 & 39.08 &  7.72 & 28.43 & 59.3 & 60.3 & 59.8\\
GPT-4o & 44.98 & 13.12 & 22.30 & 40.55 & 10.05 & 30.29 & 60.31 &  60.22 & 60.77\\
\midrule
\textbf{OCR} & \multicolumn{7}{@{}l}{} \\
\midrule
Pytesseract & 26.27 &  1.03 & 9.26 & 17.07 & 0.06 & 21.18 & 34.89 & 41.15 & 37.71 \\
MMOCR & 24.35 & 8.96 & 12.73 & 23.4 & 4.03 & 27.62 & 34.32 & 49.39 & 40.40\\
MMOCR + Llama & 28.37 & 5.37 & 15.49 & 24.94 & 2.42 & 25.0 & 52.51 & 56.88 & 54.58\\
\midrule
\textbf{Zero-Shot} & \multicolumn{7}{@{}l}{} \\
\midrule
Llama-3.2-11B-V  & 20.7 & 4.29 & 11.01 & 18.88 & 1.75 & 18.07 & 43.51 & 44.46 & 43.75\\
Qwen2-VL-7B & 20.63 & 1.93 & 12.08 & 18.97 & 0.63 & 16.13 & 46.81 & 48.35 & 47.53\\
LLaVA-NeXT     & 29.89 & 6.61 & 16.0 & 27.02 & 3.41 & 19.57 & 53.02 & 51.10 & 51.89 \\
mPLUG-DocOwl2      & 35.62 & 8.79 & 19.06 & 32.07 & 3.36 & 18.35 & 58.35 & 55.69 & 56.99 \\
MiniCPM & 39.88 & 11.11 & 20.14 & 35.45 & 7.18 & 23.76 & 59.54 & 58.91 & 59.22\\
\midrule
\textbf{Chain of Thought} & \multicolumn{7}{@{}l}{} \\
\midrule
Llama 3.2-11B-V & 20.05 & 3.4 & 10.77 & 18.14 & 1.7 & 8.57 & 42.43 & 45.89 & 43.86\\
Qwen2-VL-7B & 25.58 & 2.92 & 13.75 & 23.24 & 1.52 & 15.65 & 54.48 & 51.97 & 53.16\\
LLaVA-NeXT & 30.25 &  6.16 & 16.25 & 27.48 & 2.95 & 24.53 & 48.79 & 50.89 & 49.78 \\
mPLUG-DocOwl2      & 37.04 & 9.15 & 19.71 & 33.45 & 3.98 & 19.6 & 58.59 & 56.26 & 57.40 \\
MiniCPM & 41.50 & 11.68 & 21.04 & 37.08 & 8.60 & 26.34 & 59.32 & 58.29 & 58.80\\
\midrule
\textbf{Fine-tuning MLLMs} & \multicolumn{7}{@{}l}{} \\
\midrule
Qwen2-VL-7B & 28.77 & 6.11 & 15.18 & 26.32 & 2.66 & 19.09 & 53.78 & 51.99 & 52.83\\
LLaVA-NeXT & 31.77 & 9.94 & 18.25 & 28.7 & 6.21 & 27.18 & 51.29 & 58.89 & 54.67 \\
Llama-3.2-11B-V  & 35.16 & 13.33 & 20.64 & 31.75 & 8.91 & 28.32 & 50.65 & 58.82 & 54.19 \\
\midrule
\textsc{\textbf{Segment \& Summarize}} & \multicolumn{7}{@{}l}{} \\
\midrule
Ours  & \textbf{46.68} & \textbf{15.73} & \textbf{24.18} & \textbf{42.5} & \textbf{12.63} & \textbf{30.87} & \textbf{61.21} & \textbf{61.62} & \textbf{61.37} \\
\bottomrule
\end{tabular}
}
\caption{Summarization results on the \textsc{PosterSum} dataset. The results show ROUGE scores (R-1, R-2, R-L, R-LSum), BERTScores (BS\textsubscript{p}, BS\textsubscript{r}, BS\textsubscript{f1}), SacreBLEU, and METEOR scores for all the baseline and models.  All the scores are percentages.}
\label{tab:summ_result}
\end{table*}
\begin{figure*}[tb]
  \centering
  \includegraphics[width={\textwidth}]{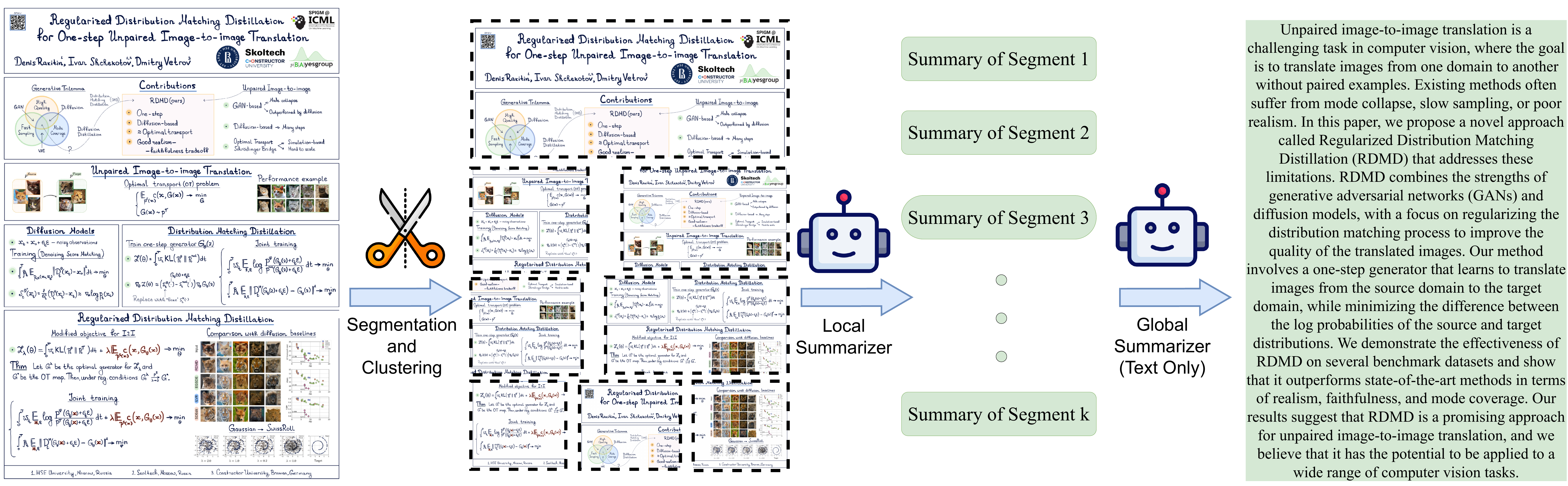}
  \caption{Illustration of our \method pipeline. The poster, describing the work in \citet{rakitin2024regularized}, is first divided into segments, each of which is summarized by a MLLM. These localized summaries are subsequently merged by a text-based large language model to generate a single, coherent summary.
  }
  \label{fig:segment_summ}
\end{figure*}

\subsection{\method}
We now introduce \method, a hierarchical approach inspired by the divide-and-conquer principle.
Rather than processing the entire poster $I$ as a single input, \method decomposes the task into three key steps: \begin{inparaenum} \item Segmentation and Clustering \item Localized Summarization, and \item Global Summarization. \end{inparaenum}
The \method pipeline is outlined in \cref{fig:segment_summ}.

\paragraph{1. Segmentation and Clustering.} Given the image of a poster \( I \), the first step is to segment it into \( n \) coherent regions \( M = \{M_1, M_2, \dots, M_n\} \).
This is achieved using a segmentation model \( S_\phi \), parameterized by \( \phi \),
Since the number of regions \( n \) can be large and can contain redundant and small segments, the regions are further clustered into groups \(R\) with the number of the clustered regions as \( k \)  using a clustering algorithm \( C\) such that  \( k \ll n \). The clustering step groups similar regions together, reducing redundancy and ensuring complete coverage of the poster. Formally, $M = S_\phi(I)$ and  $R = C(M)$.

By segmenting the poster and summarizing each region independently, the method ensures a detailed and accurate understanding of the content.

\paragraph{2. Localized Summarization.} For each clustered region \( R_i \), a localized summary \( \hat{Y}_i = \{\hat{y}_{i1}, \hat{y}_{i2}, \dots, \hat{y}_{ik}\} \) is generated using an MLLM~$V_\phi$. The model is used to extract and interpret the content within \( R_i \), including text, figures, and tables, to generate a localized summary for that specific region. This also helps in processing the high-resolution image.

\paragraph{3. Global Summarization.} The localized summaries \( \hat{Y}_1, \hat{Y}_2, \dots, \hat{Y}_k \) are combined into a cohesive global summary \( \hat{Y} \) using a text-based large language model \( L_\omega\), parameterized by \( \omega \).
The model \( L_\omega \) takes as input the individual summaries and generates a single, well-structured output that represents the overall content of the poster.
This step ensures that the final abstract is not only comprehensive but also maintains logical flow and coherence. Formally, $\hat{Y} = L_\omega(\hat{Y}_1, \hat{Y}_2, \dots, \hat{Y}_k)$.

This processing pipeline helps summary generation through a structured, localized, and hierarchical approach. By segmenting the poster and summarizing each region independently, the method captures fine-grained details that might be overlooked in a global approach. This also aligns with the structure of these posters, which are mostly divided into sections. This approach does not require additional training or fine-tuning, and both the models ($V_\phi$, $L_\omega$) are frozen.

\section{Experimental Details}
All the models in each category were evaluated using the same hyperparameter settings for fair evaluation.
We generate at most 768 new tokens for all the experiments.
For closed-source models, we used the default platform settings. 
Open-source models were evaluated with a beam size of 4 with greedy decoding to ensure reproducibility.
%
%
The fine-tuning experiments were conducted for $10$ epochs with a batch size of $4$.
More details about the hyperparameters and prompt templates can be found in \cref{sec:templates,experiment_details_extra}.

For \method, we used the Segment Anything Model~\citep{DBLP:conf/iccv/KirillovMRMRGXW23} for segmentation with k-Means for clustering. The number of clusters ($k$) was set to 8 based on the analysis in \cref{sec:number_of_cluster}.
We used MiniCPM-Llama3-V-2.5 as the local summarize ($V_\phi$) and  Llama 3.1-8B-Instruct as the global summarizer ($L_\omega$).
We used the training set for fine-tuning and the validation set for hyperparameter tuning. All the final results are evaluated on the test set.

\paragraph{Evaluation Metrics.}
We use ROUGE F1 (R-1/2/L/LSum) scores~\citep{lin-2004-rouge}, SacreBLEU~\cite[SBLEU;][]{post-2018-call}, METEOR~\cite[MET;][]{banerjee-lavie-2005-meteor}, and BERTScore~\citep{bert-score} to evaluate the accuracy of 
all models. 

\section{Results}
\label{sec:results}

\cref{tab:summ_result} presents the poster summarization performance of all baselines alongside our proposed \method\ method, evaluated on the \textsc{PosterSum} test set. Our method outperforms both open-source and closed-source models, achieving the best results across all metrics.

\paragraph{Closed-source Models.}
GPT-4o achieves relatively high performance among the closed-source models across all metrics, with ROUGE-1/2/L scores of 44.98, 13.12, and 22.30, respectively. Claude-3.5 Sonnet also performs well, attaining a ROUGE-L score of 19.51.

\paragraph{OCR Baselines.}
The two OCR-based methods, MMOCR and Pytesseract, achieve relatively low scores across all metrics. This is likely due to the limitation of concatenating raw OCR text without leveraging other visual elements. Combining OCR with the text-only Llama-3.1 model results in a substantial improvement, with ROUGE-L, increasing from 12.73 to 15.49. Interestingly, these OCR methods still outperform certain multimodal models, indicating that text extraction remains a challenge for some MLLMs.

\paragraph{Open-source Models.}
Among the open-source MLLMs evaluated in zero-shot settings, MiniCPM-Llama3-V-2.5 obtains the highest ROUGE-1/L score (39.88/20.14) and a strong BERTScore-F1 of 59.22. Meanwhile, mPLUG-DocOwl2 achieves a competitive ROUGE-L of 19.06 and a BERTScore-F1 of 56.99.

\paragraph{Chain of Thought (CoT).}
Adding an explicit CoT prompt improves the performance of most models. For instance, MiniCPM-Llama3-V-2.5 improves its ROUGE-1/L/METEOR scores to 41.50/21.04/26.34, while mPLUG-DocOwl2's performance also increases (ROUGE-1/L of 37.04/19.71). Additionally, LLaVA-NeXT and Qwen2-VL-7B exhibit similar gains. Although the performance boosts are not large, these results suggest that guiding models via CoT prompt can help in extracting relevant poster content.

\paragraph{Fine-tuned Models.}
Using LoRA substantially boosts performance for both MLLMs. In particular, Llama-3.2-11B-Instruct demonstrates notable improvements in ROUGE, ScareBLEU, and METEOR scores, though it does not surpass the best CoT variants of \textsc{mPLUG-DocOwl2} and \textsc{MiniCPM-Llama3-V-2.5}, which likely benefit from pre-training on multimodal scientific data.

\paragraph{\method.}
Our proposed method outperforms all other models, including closed-source models, on all metrics, achieving ROUGE-1/2/L scores of 46.68, 15.73, and 24.18, respectively, with a 3.14\% gain on ROUGE-L compared to open-source models. It also attains a substantially higher ScareBLEU score (12.63) and a BERTScore-F1 of 61.37. These results indicate that local-region summaries effectively preserve small details and handle posters of varying complexity by processing each region independently rather than attempting to analyze the entire poster as a single input.
\begin{table}[t]
\centering
\scalebox{0.9}{
\begin{tabular}{lcccc}
\toprule
\textbf{Methods} & \textbf{R1} & \textbf{R-2} & \textbf{R-L} & \textbf{MET}\\
\midrule
Without clustering & 42.25 & 14.30 & 22.76 & 23.97 \\
With clustering &  46.68 & 15.73 & 24.18 &  30.87\\
\bottomrule
\end{tabular}
}
\caption{Comparison of \method with and without clustering --- clustering the segments yields more accurate results.}
\label{tab:clustering_ablation}
\end{table}
\begin{table}[t]
\centering
\scalebox{0.85}{
\begin{tabular}{lcccc}
\toprule
\textbf{Methods} & \textbf{R1} & \textbf{R-2} & \textbf{R-L} & \textbf{MET}\\
\midrule
mPLUG-DocOwl2 & 37.04 & 9.15 & 19.71 & 19.6 \\
Ours with DocOwl2 & 42.48 & 11.18 & 20.61 &  26.72\\
Ours with MiniCPM &  46.68 & 15.73 & 24.18 &  30.87\\
\bottomrule
\end{tabular}
}
\caption{Comparison of using mPLUG-DocOwl2 as local summarize. Applying \method\ shows improvement compared to using the model itself.}
\label{tab:local_summarization}
\end{table}
\section{Ablation Studies and Analysis}
\paragraph{Effect of Clustering on Summarization.}
To quantify the impact of clustering in our \method\ approach, we conduct an ablation study that removes the clustering step.
Specifically, we select the top-$k$ segments (with $k=8$) based on their region size to generate local and global summaries. \cref{tab:clustering_ablation} shows that clustering improves the ROUGE-1 score by +4.43, ROUGE-2 by +1.43, and ROUGE-L by +1.42 over the non-clustered baseline.
We hypothesize that clustering helps reduce redundant segments and improves context aggregation.

\paragraph{Effect of Local Vision Summarization.}
To assess the role of the local summarization model in \method, we replaced MiniCPM-Llama3-V-2.5 with mPLUG-DocOwl2, which previously ranked second among open-source models under the CoT setting. \cref{tab:local_summarization} shows that using mPLUG-DocOwl2 with our hierarchical approach boosts ROUGE-1 to 42.48 and METEOR to 26.72 compared to using the model in the CoT setting. However, it does not outperform our method using MiniCPM. These findings highlight that the segmentation and summarization approach substantially improves performance compared to using the poster as a single input.  

\paragraph{Challenges in Human Evaluation and Reliance on Automatic Metrics}
Evaluating scientific summaries against their posters is both costly and logistically complex for human annotators. Scientific posters consist of dense technical content (including specialized terminology, tables, figures, and equations), requiring domain expertise and making the recruitment of qualified annotators time-consuming and expensive. Moreover, the diversity of research topics could lead to inconsistent judgments even among experts.
For this reason, we rely on automatic metrics. Additionally, we conducted a factuality evaluation, as discussed in \cref{sec:factuality}. However, existing factuality metrics, such as SummaC Conv~\citep{laban-etal-2022-summac} and FActScore~\citep{min-etal-2023-factscore}, perform poorly on scientific text, highlighting the need for improved evaluation methods for multimodal scientific data.

\section{Conclusions}
We presented \textsc{PosterSum}, a multimodal benchmark for scientific poster summarization comprising 16,305 poster-abstract pairs.
Our experiments show that even state-of-the-art MLLMs struggle with key aspects of scientific poster summarization.
Furthermore, we propose \method, a hierarchical approach that outperforms existing models by breaking down the summarization task into localized segments before generating a cohesive abstract.
We find that our method outperforms MLLMs in both zero-shot and fine-tuned settings and that there remains significant room for improvement in multimodal understanding of complex scientific documents such as posters.
We believe \textsc{PosterSum} will be a valuable resource for developing and evaluating MLLMs capable of processing information-dense scientific content.
\section*{Acknowledgments}
%
This work was supported in part by the School of Informatics at the University of Edinburgh.
Pasquale Minervini was partially funded by ELIAI (The Edinburgh Laboratory for Integrated Artificial Intelligence), EPSRC (grant no.\ EP/W002876/1), and a donation from Accenture LLP.
This work was supported by the Edinburgh International Data Facility (EIDF) and the Data-Driven Innovation Programme at the University of Edinburgh.

\section*{Limitations}
While our work advances scientific poster summarization, we should highlight a few limitations.
First, our dataset is restricted to machine learning conference posters from 2022 to 2024, which may limit the generalization to other scientific domains.
Second, while practical, automated topic labeling using GPT-4o may introduce biases or inaccuracies in the topic distribution.
The proposed \method\ method relies heavily on the quality of the initial segmentation --- suboptimal segmentation can lead to fragmented or redundant local summaries.
Our method also assumes that the content can be meaningfully decomposed into spatial regions, which may not hold for posters with complex cross-referencing or interdependent visual elements.
We considered the abstract as a ground-truth summary of the poster, but the poster may sometimes differ from the paper.

\section*{Ethics Statement}
\paragraph{Dataset.} All the scientific posters and abstracts in our dataset are sourced from publicly accessible conference resources. Additionally, we sought permission from the conference website contacts to use the publicly available data for research purposes.

\paragraph{Multimodal Large Language Models.} This paper utilizes pre-trained multimodal large language models, which have been shown to exhibit various biases, occasionally hallucinate, and generate non-faithful text. Therefore, summaries generated using our dataset should not be released without automatic filtering or manual verification to ensure accuracy and reliability.

\paragraph{Bias.} Despite efforts to include a wide range of posters, the dataset may not fully represent the diversity of research poster styles, languages, or scientific disciplines. As a result, models trained on \textsc{PosterSum} may exhibit biases towards the types of posters included in the dataset. Future work should consider expanding the dataset to encompass a broader spectrum of academic fields and visual formats to mitigate potential biases.
\bibliography{acl_latex}

\begin{thebibliography}{55}
\providecommand{\natexlab}[1]{#1}

\bibitem[{Abreu et~al.(2022)Abreu, Vaska, and Helus}]{example2_paper}
Natalie Abreu, Nathan Vaska, and Victoria Helus. 2022.
\newblock \href {https://doi.org/10.48550/ARXIV.2211.11880} {Addressing mistake severity in neural networks with semantic knowledge}.
\newblock \emph{CoRR}, abs/2211.11880.

\bibitem[{Alayrac et~al.(2022)Alayrac, Donahue, Luc, Miech, Barr, Hasson, Lenc, Mensch, Millican, Reynolds, Ring, Rutherford, Cabi, Han, Gong, Samangooei, Monteiro, Menick, Borgeaud, Brock, Nematzadeh, Sharifzadeh, Binkowski, Barreira, Vinyals, Zisserman, and Simonyan}]{DBLP:conf/nips/AlayracDLMBHLMM22}
Jean{-}Baptiste Alayrac, Jeff Donahue, Pauline Luc, Antoine Miech, Iain Barr, Yana Hasson, Karel Lenc, Arthur Mensch, Katherine Millican, Malcolm Reynolds, Roman Ring, Eliza Rutherford, Serkan Cabi, Tengda Han, Zhitao Gong, Sina Samangooei, Marianne Monteiro, Jacob~L. Menick, Sebastian Borgeaud, and 8 others. 2022.
\newblock \href {https://openreview.net/forum?id=EbMuimAbPbs} {Flamingo: a visual language model for few-shot learning}.
\newblock In \emph{NeurIPS}.

\bibitem[{Anil et~al.(2024)Anil, Borgeaud, Alayrac, Yu, Soricut, Schalkwyk, Dai, Hauth, Millican, Silver, Johnson, Antonoglou, Schrittwieser, Glaese, Chen, Pitler, Lillicrap, Lazaridou, Firat, Molloy, Isard, Barham, Hennigan, Lee, Viola, Reynolds, Xu, Doherty, Collins, Meyer, Rutherford, Moreira, Ayoub, Goel, Krawczyk, Du, Chi, Cheng, Ni, Shah, Kane, Chan, Faruqui, Severyn, Lin, Li, Cheng, Ittycheriah, Mahdieh, Chen, Sun, Tran, Bagri, Lakshminarayanan, Liu, Orban, Güra, Zhou, Song, Boffy, Ganapathy, Zheng, Choe, Ágoston Weisz, Zhu, Lu, Gopal, Kahn, Kula, Pitman, Shah, Taropa, Merey, Baeuml, Chen, Shafey, Zhang, Sercinoglu, Tucker, Piqueras, Krikun, Barr, Savinov, Danihelka, Roelofs, White, Andreassen, von Glehn, Yagati, Kazemi, Gonzalez, Khalman, Sygnowski, Frechette, Smith, Culp, Proleev, Luan, Chen, Lottes, Schucher, Lebron, Rrustemi, Clay, Crone, Kocisky, Zhao, Perz, Yu, Howard, Bloniarz, Rae, Lu, Sifre, Maggioni, Alcober, Garrette, Barnes, Thakoor, Austin, Barth-Maron, Wong, Joshi, Chaabouni, Fatiha,
  Ahuja, Tomar, Senter, Chadwick, Kornakov, Attaluri, Iturrate, Liu, Li, Cogan, Chen, Jia, Gu, Zhang, Grimstad, Hartman, Garcia, Pillai, Devlin, Laskin, de~Las~Casas, Valter, Tao, Blanco, Badia, Reitter, Chen, Brennan, Rivera, Brin, Iqbal, Surita, Labanowski, Rao, Winkler, Parisotto, Gu, Olszewska, Addanki, Miech, Louis, Teplyashin, Brown, Catt, Balaguer, Xiang, Wang, Ashwood, Briukhov, Webson, Ganapathy, Sanghavi, Kannan, Chang, Stjerngren, Djolonga, Sun, Bapna, Aitchison, Pejman, Michalewski, Yu, Wang, Love, Ahn, Bloxwich, Han, Humphreys, Sellam, Bradbury, Godbole, Samangooei, Damoc, Kaskasoli, Arnold, Vasudevan, Agrawal, Riesa, Lepikhin, Tanburn, Srinivasan, Lim, Hodkinson, Shyam, Ferret, Hand, Garg, Paine, Li, Li, Giang, Neitz, Abbas, York, Reid, Cole, Chowdhery, Das, Rogozińska, Nikolaev, Sprechmann, Nado, Zilka, Prost, He, Monteiro, Mishra, Welty, Newlan, Jia, Allamanis, Hu, de~Liedekerke, Gilmer, Saroufim, Rijhwani, Hou, Shrivastava, Baddepudi, Goldin, Ozturel, Cassirer, Xu, Sohn, Sachan, Amplayo,
  Swanson, Petrova, Narayan, Guez, Brahma, Landon, Patel, Zhao, Villela, Wang, Jia, Rahtz, Giménez, Yeung, Keeling, Georgiev, Mincu, Wu, Haykal, Saputro, Vodrahalli, Qin, Cankara, Sharma, Fernando, Hawkins, Neyshabur, Kim, Hutter, Agrawal, Castro-Ros, van~den Driessche, Wang, Yang, yiin Chang, Komarek, McIlroy, Lučić, Zhang, Farhan, Sharman, Natsev, Michel, Bansal, Qiao, Cao, Shakeri, Butterfield, Chung, Rubenstein, Agrawal, Mensch, Soparkar, Lenc, Chung, Pope, Maggiore, Kay, Jhakra, Wang, Maynez, Phuong, Tobin, Tacchetti, Trebacz, Robinson, Katariya, Riedel, Bailey, Xiao, Ghelani, Aroyo, Slone, Houlsby, Xiong, Yang, Gribovskaya, Adler, Wirth, Lee, Li, Kagohara, Pavagadhi, Bridgers, Bortsova, Ghemawat, Ahmed, Liu, Powell, Bolina, Iinuma, Zablotskaia, Besley, Chung, Dozat, Comanescu, Si, Greer, Su, Polacek, Kaufman, Tokumine, Hu, Buchatskaya, Miao, Elhawaty, Siddhant, Tomasev, Xing, Greer, Miller, Ashraf, Roy, Zhang, Ma, Filos, Besta, Blevins, Klimenko, Yeh, Changpinyo, Mu, Chang, Pajarskas, Muir, Cohen,
  Lan, Haridasan, Marathe, Hansen, Douglas, Samuel, Wang, Austin, Lan, Jiang, Chiu, Lorenzo, Sjösund, Cevey, Gleicher, Avrahami, Boral, Srinivasan, Selo, May, Aisopos, Hussenot, Soares, Baumli, Chang, Recasens, Caine, Pritzel, Pavetic, Pardo, Gergely, Frye, Ramasesh, Horgan, Badola, Kassner, Roy, Dyer, Campos, Tomala, Tang, Badawy, White, Mustafa, Lang, Jindal, Vikram, Gong, Caelles, Hemsley, Thornton, Feng, Stokowiec, Zheng, Thacker, Çağlar Ünlü, Zhang, Saleh, Svensson, Bileschi, Patil, Anand, Ring, Tsihlas, Vezer, Selvi, Shevlane, Rodriguez, Kwiatkowski, Daruki, Rong, Dafoe, FitzGerald, Gu-Lemberg, Khan, Hendricks, Pellat, Feinberg, Cobon-Kerr, Sainath, Rauh, Hashemi, Ives, Hasson, Noland, Cao, Byrd, Hou, Wang, Sottiaux, Paganini, Lespiau, Moufarek, Hassan, Shivakumar, van Amersfoort, Mandhane, Joshi, Goyal, Tung, Brock, Sheahan, Misra, Li, Rakićević, Dehghani, Liu, Mittal, Oh, Noury, Sezener, Huot, Lamm, Cao, Chen, Mudgal, Stella, Brooks, Vasudevan, Liu, Chain, Melinkeri, Cohen, Wang, Seymore,
  Zubkov, Goel, Yue, Krishnakumaran, Albert, Hurley, Sano, Mohananey, Joughin, Filonov, Kępa, Eldawy, Lim, Rishi, Badiezadegan, Bos, Chang, Jain, Padmanabhan, Puttagunta, Krishna, Baker, Kalb, Bedapudi, Kurzrok, Lei, Yu, Litvin, Zhou, Wu, Sobell, Siciliano, Papir, Neale, Bragagnolo, Toor, Chen, Anklin, Wang, Feng, Gholami, Ling, Liu, Walter, Moghaddam, Kishore, Adamek, Mercado, Mallinson, Wandekar, Cagle, Ofek, Garrido, Lombriser, Mukha, Sun, Mohammad, Matak, Qian, Peswani, Janus, Yuan, Schelin, David, Garg, He, Duzhyi, Älgmyr, Lottaz, Li, Yadav, Xu, Chinien, Shivanna, Chuklin, Li, Spadine, Wolfe, Mohamed, Das, Dai, He, von Dincklage, Upadhyay, Maurya, Chi, Krause, Salama, Rabinovitch, M, Selvan, Dektiarev, Ghiasi, Guven, Gupta, Liu, Sharma, Shtacher, Paul, Akerlund, Aubet, Huang, Zhu, Zhu, Teixeira, Fritze, Bertolini, Marinescu, Bölle, Paulus, Gupta, Latkar, Chang, Sanders, Wilson, Wu, Tan, Thiet, Doshi, Lall, Mishra, Chen, Luong, Benjamin, Lee, Andrejczuk, Rabiej, Ranjan, Styrc, Yin, Simon, Harriott,
  Bansal, Robsky, Bacon, Greene, Mirylenka, Zhou, Sarvana, Goyal, Andermatt, Siegler, Horn, Israel, Pongetti, Chen, Selvatici, Silva, Wang, Tolins, Guu, Yogev, Cai, Agostini, Shah, Nguyen, Donnaile, Pereira, Friso, Stambler, Kurzrok, Kuang, Romanikhin, Geller, Yan, Jang, Lee, Fica, Malmi, Tan, Banica, Balle, Pham, Huang, Avram, Shi, Singh, Hidey, Ahuja, Saxena, Dooley, Potharaju, O'Neill, Gokulchandran, Foley, Zhao, Dusenberry, Liu, Mehta, Kotikalapudi, Safranek-Shrader, Goodman, Kessinger, Globen, Kolhar, Gorgolewski, Ibrahim, Song, Eichenbaum, Brovelli, Potluri, Lahoti, Baetu, Ghorbani, Chen, Crawford, Pal, Sridhar, Gurita, Mujika, Petrovski, Cedoz, Li, Chen, Santo, Goyal, Punjabi, Kappaganthu, Kwak, LV, Velury, Choudhury, Hall, Shah, Figueira, Thomas, Lu, Zhou, Kumar, Jurdi, Chikkerur, Ma, Yu, Kwak, Ähdel, Rajayogam, Choma, Liu, Barua, Ji, Park, Hellendoorn, Bailey, Bilal, Zhou, Khatir, Sutton, Rzadkowski, Macintosh, Shagin, Medina, Liang, Zhou, Shah, Bi, Dankovics, Banga, Lehmann, Bredesen, Lin,
  Hoffmann, Lai, Chung, Yang, Balani, Bražinskas, Sozanschi, Hayes, Alcalde, Makarov, Chen, Stella, Snijders, Mandl, Kärrman, Nowak, Wu, Dyck, Vaidyanathan, R, Mallet, Rudominer, Johnston, Mittal, Udathu, Christensen, Verma, Irving, Santucci, Elsayed, Davoodi, Georgiev, Tenney, Hua, Cideron, Leurent, Alnahlawi, Georgescu, Wei, Zheng, Scandinaro, Jiang, Snoek, Sundararajan, Wang, Ontiveros, Karo, Cole, Rajashekhar, Tumeh, Ben-David, Jain, Uesato, Datta, Bunyan, Wu, Zhang, Stanczyk, Zhang, Steiner, Naskar, Azzam, Johnson, Paszke, Chiu, Elias, Mohiuddin, Muhammad, Miao, Lee, Vieillard, Park, Zhang, Stanway, Garmon, Karmarkar, Dong, Lee, Kumar, Zhou, Evens, Isaac, Irving, Loper, Fink, Arkatkar, Chen, Shafran, Petrychenko, Chen, Jia, Levskaya, Zhu, Grabowski, Mao, Magni, Yao, Snaider, Casagrande, Palmer, Suganthan, Castaño, Giannoumis, Kim, Rybiński, Sreevatsa, Prendki, Soergel, Goedeckemeyer, Gierke, Jafari, Gaba, Wiesner, Wright, Wei, Vashisht, Kulizhskaya, Hoover, Le, Li, Iwuanyanwu, Liu, Ramirez, Khorlin,
  Cui, LIN, Wu, Aguilar, Pallo, Chakladar, Perng, Abellan, Zhang, Dasgupta, Kushman, Penchev, Repina, Wu, van~der Weide, Ponnapalli, Kaplan, Simsa, Li, Dousse, Yang, Piper, Ie, Pasumarthi, Lintz, Vijayakumar, Andor, Valenzuela, Lui, Paduraru, Peng, Lee, Zhang, Greene, Nguyen, Kurylowicz, Hardin, Dixon, Janzer, Choo, Feng, Zhang, Singhal, Du, McKinnon, Antropova, Bolukbasi, Keller, Reid, Finchelstein, Raad, Crocker, Hawkins, Dadashi, Gaffney, Franko, Bulanova, Leblond, Chung, Askham, Cobo, Xu, Fischer, Xu, Sorokin, Alberti, Lin, Evans, Dimitriev, Forbes, Banarse, Tung, Omernick, Bishop, Sterneck, Jain, Xia, Amid, Piccinno, Wang, Banzal, Mankowitz, Polozov, Krakovna, Brown, Bateni, Duan, Firoiu, Thotakuri, Natan, Geist, tan Girgin, Li, Ye, Roval, Tojo, Kwong, Lee-Thorp, Yew, Sinopalnikov, Ramos, Mellor, Sharma, Wu, Miller, Sonnerat, Vnukov, Greig, Beattie, Caveness, Bai, Eisenschlos, Korchemniy, Tsai, Jasarevic, Kong, Dao, Zheng, Liu, Yang, Zhu, Teh, Sanmiya, Gladchenko, Trdin, Toyama, Rosen, Tavakkol, Xue,
  Elkind, Woodman, Carpenter, Papamakarios, Kemp, Kafle, Grunina, Sinha, Talbert, Wu, Owusu-Afriyie, Du, Thornton, Pont-Tuset, Narayana, Li, Fatehi, Wieting, Ajmeri, Uria, Ko, Knight, Héliou, Niu, Gu, Pang, Li, Levine, Stolovich, Santamaria-Fernandez, Goenka, Yustalim, Strudel, Elqursh, Deck, Lee, Li, Levin, Hoffmann, Holtmann-Rice, Bachem, Arora, Koh, Yeganeh, Põder, Tariq, Sun, Ionita, Seyedhosseini, Tafti, Liu, Gulati, Liu, Ye, Chrzaszcz, Wang, Sethi, Li, Brown, Singh, Fan, Parisi, Stanton, Koverkathu, Choquette-Choo, Li, Lu, Ittycheriah, Shroff, Varadarajan, Bahargam, Willoughby, Gaddy, Desjardins, Cornero, Robenek, Mittal, Albrecht, Shenoy, Moiseev, Jacobsson, Ghaffarkhah, Rivière, Walton, Crepy, Parrish, Zhou, Farabet, Radebaugh, Srinivasan, van~der Salm, Fidjeland, Scellato, Latorre-Chimoto, Klimczak-Plucińska, Bridson, de~Cesare, Hudson, Mendolicchio, Walker, Morris, Mauger, Guseynov, Reid, Odoom, Loher, Cotruta, Yenugula, Grewe, Petrushkina, Duerig, Sanchez, Yadlowsky, Shen, Globerson, Webb, Dua,
  Li, Bhupatiraju, Hurt, Qureshi, Agarwal, Shani, Eyal, Khare, Belle, Wang, Tekur, Kale, Wei, Sang, Saeta, Liechty, Sun, Zhao, Lee, Nayak, Fritz, Vuyyuru, Aslanides, Vyas, Wicke, Ma, Eltyshev, Martin, Cate, Manyika, Amiri, Kim, Xiong, Kang, Luisier, Tripuraneni, Madras, Guo, Waters, Wang, Ainslie, Baldridge, Zhang, Pruthi, Bauer, Yang, Mansour, Gelman, Xu, Polovets, Liu, Cai, Chen, Sheng, Xue, Ozair, Angermueller, Li, Sinha, Wang, Wiesinger, Koukoumidis, Tian, Iyer, Gurumurthy, Goldenson, Shah, Blake, Yu, Urbanowicz, Palomaki, Fernando, Durden, Mehta, Momchev, Rahimtoroghi, Georgaki, Raul, Ruder, Redshaw, Lee, Zhou, Jalan, Li, Hechtman, Schuh, Nasr, Milan, Mikulik, Franco, Green, Nguyen, Kelley, Mahendru, Hu, Howland, Vargas, Hui, Bansal, Rao, Ghiya, Wang, Ye, Sarr, Preston, Elish, Li, Kaku, Gupta, Pasupat, Juan, Someswar, M., Chen, Amini, Fabrikant, Chu, Dong, Muthal, Buthpitiya, Jauhari, Hua, Khandelwal, Hitron, Ren, Rinaldi, Drath, Dabush, Jiang, Godhia, Sachs, Chen, Fan, Taitelbaum, Noga, Dai, Wang,
  Liang, Hamer, Ferng, Elkind, Atias, Lee, Listík, Carlen, van~de Kerkhof, Pikus, Zaher, Müller, Zykova, Stefanec, Gatsko, Hirnschall, Sethi, Xu, Ahuja, Tsai, Stefanoiu, Feng, Dhandhania, Katyal, Gupta, Parulekar, Pitta, Zhao, Bhatia, Bhavnani, Alhadlaq, Li, Danenberg, Tu, Pine, Filippova, Ghosh, Limonchik, Urala, Lanka, Clive, Sun, Li, Wu, Hongtongsak, Li, Thakkar, Omarov, Majmundar, Alverson, Kucharski, Patel, Jain, Zabelin, Pelagatti, Kohli, Kumar, Kim, Sankar, Shah, Ramachandruni, Zeng, Bariach, Weidinger, Vu, Andreev, He, Hui, Kashem, Subramanya, Hsiao, Hassabis, Kavukcuoglu, Sadovsky, Le, Strohman, Wu, Petrov, Dean, and Vinyals}]{geminiteam2024}
Rohan Anil, Sebastian Borgeaud, Jean-Baptiste Alayrac, Jiahui Yu, Radu Soricut, Johan Schalkwyk, Andrew~M. Dai, Anja Hauth, Katie Millican, David Silver, Melvin Johnson, Ioannis Antonoglou, Julian Schrittwieser, Amelia Glaese, Jilin Chen, Emily Pitler, Timothy Lillicrap, Angeliki Lazaridou, Orhan Firat, and 1330 others. 2024.
\newblock \href {https://arxiv.org/abs/2312.11805} {Gemini: A family of highly capable multimodal models}.
\newblock \emph{Preprint}, arXiv:2312.11805.

\bibitem[{Anthropic(2024)}]{anthropic_claude3_5_sonnet}
Anthropic. 2024.
\newblock Claude 3.5 - sonnet.
\newblock \url{https://www.anthropic.com/news/claude-3-5-sonnet}.
\newblock Accessed: 2024-12-06.

\bibitem[{Appalaraju et~al.(2024)Appalaraju, Tang, Dong, Sankaran, Zhou, and Manmatha}]{DocFormerv2}
Srikar Appalaraju, Peng Tang, Qi~Dong, Nishant Sankaran, Yichu Zhou, and R.~Manmatha. 2024.
\newblock \href {https://doi.org/10.1609/AAAI.V38I2.27828} {Docformerv2: Local features for document understanding}.
\newblock In \emph{Thirty-Eighth {AAAI} Conference on Artificial Intelligence, {AAAI} 2024, Thirty-Sixth Conference on Innovative Applications of Artificial Intelligence, {IAAI} 2024, Fourteenth Symposium on Educational Advances in Artificial Intelligence, {EAAI} 2014, February 20-27, 2024, Vancouver, Canada}, pages 709--718. {AAAI} Press.

\bibitem[{Banerjee and Lavie(2005)}]{banerjee-lavie-2005-meteor}
Satanjeev Banerjee and Alon Lavie. 2005.
\newblock \href {https://aclanthology.org/W05-0909/} {{METEOR}: An automatic metric for {MT} evaluation with improved correlation with human judgments}.
\newblock In \emph{Proceedings of the {ACL} Workshop on Intrinsic and Extrinsic Evaluation Measures for Machine Translation and/or Summarization}, pages 65--72, Ann Arbor, Michigan. Association for Computational Linguistics.

\bibitem[{Cachola et~al.(2020)Cachola, Lo, Cohan, and Weld}]{DBLP:conf/emnlp/CacholaLCW20}
Isabel Cachola, Kyle Lo, Arman Cohan, and Daniel Weld. 2020.
\newblock \href {https://doi.org/10.18653/v1/2020.findings-emnlp.428} {{TLDR}: Extreme summarization of scientific documents}.
\newblock In \emph{Findings of the Association for Computational Linguistics: EMNLP 2020}, pages 4766--4777, Online. Association for Computational Linguistics.

\bibitem[{Chen et~al.(2022)Chen, Tang, Liu, Zhao, Huang, and Yu}]{example1_paper}
Chaoqi Chen, Luyao Tang, Feng Liu, Gangming Zhao, Yue Huang, and Yizhou Yu. 2022.
\newblock \href {https://openreview.net/forum?id=V0GwAmDclY} {Mix and reason: Reasoning over semantic topology with data mixing for domain generalization}.
\newblock In \emph{Advances in Neural Information Processing Systems}.

\bibitem[{Chen and Zhao(2023)}]{divide_and_conquer_cvpr}
Shi Chen and Qi~Zhao. 2023.
\newblock \href {https://doi.org/10.1109/CVPR52729.2023.00651} {Divide and conquer: Answering questions with object factorization and compositional reasoning}.
\newblock In \emph{{IEEE/CVF} Conference on Computer Vision and Pattern Recognition, {CVPR} 2023, Vancouver, BC, Canada, June 17-24, 2023}, pages 6736--6745. {IEEE}.

\bibitem[{Chen et~al.(2024)Chen, Liu, Yu, Sun, Liu, Wu, Zhang, Wang, and Wang}]{m3av}
Zhe Chen, Heyang Liu, Wenyi Yu, Guangzhi Sun, Hongcheng Liu, Ji~Wu, Chao Zhang, Yu~Wang, and Yanfeng Wang. 2024.
\newblock \href {https://doi.org/10.18653/V1/2024.ACL-LONG.489} {M{\({^3}\)}av: {A} multimodal, multigenre, and multipurpose audio-visual academic lecture dataset}.
\newblock In \emph{Proceedings of the 62nd Annual Meeting of the Association for Computational Linguistics (Volume 1: Long Papers), {ACL} 2024, Bangkok, Thailand, August 11-16, 2024}, pages 9041--9060. Association for Computational Linguistics.

\bibitem[{Fabbri et~al.(2022)Fabbri, Wu, Liu, and Xiong}]{fabbri-etal-2022-qafacteval}
Alexander Fabbri, Chien-Sheng Wu, Wenhao Liu, and Caiming Xiong. 2022.
\newblock \href {https://doi.org/10.18653/v1/2022.naacl-main.187} {{QAF}act{E}val: Improved {QA}-based factual consistency evaluation for summarization}.
\newblock In \emph{Proceedings of the 2022 Conference of the North American Chapter of the Association for Computational Linguistics: Human Language Technologies}, pages 2587--2601, Seattle, United States. Association for Computational Linguistics.

\bibitem[{Fu et~al.(2024)Fu, Chen, Shen, Qin, Zhang, Lin, Yang, Zheng, Li, Sun, Wu, and Ji}]{fu2023_MME}
Chaoyou Fu, Peixian Chen, Yunhang Shen, Yulei Qin, Mengdan Zhang, Xu~Lin, Jinrui Yang, Xiawu Zheng, Ke~Li, Xing Sun, Yunsheng Wu, and Rongrong Ji. 2024.
\newblock \href {https://arxiv.org/abs/2306.13394} {Mme: A comprehensive evaluation benchmark for multimodal large language models}.
\newblock \emph{Preprint}, arXiv:2306.13394.

\bibitem[{Garg et~al.(2024)Garg, Burns, Karagol~Ayan, Bitton, Montgomery, Onoe, Bunner, Krishna, Baldridge, and Soricut}]{garg-etal-2024-imageinwords}
Roopal Garg, Andrea Burns, Burcu Karagol~Ayan, Yonatan Bitton, Ceslee Montgomery, Yasumasa Onoe, Andrew Bunner, Ranjay Krishna, Jason~Michael Baldridge, and Radu Soricut. 2024.
\newblock \href {https://doi.org/10.18653/v1/2024.emnlp-main.6} {{I}mage{I}n{W}ords: Unlocking hyper-detailed image descriptions}.
\newblock In \emph{Proceedings of the 2024 Conference on Empirical Methods in Natural Language Processing}, pages 93--127, Miami, Florida, USA. Association for Computational Linguistics.

\bibitem[{Grattafiori et~al.(2024)Grattafiori, Dubey, Jauhri, Pandey, and et~al.}]{llama3herd}
Aaron Grattafiori, Abhimanyu Dubey, Abhinav Jauhri, Abhinav Pandey, and Abhishek~Kadian et~al. 2024.
\newblock \href {https://arxiv.org/abs/2407.21783} {The llama 3 herd of models}.
\newblock \emph{Preprint}, arXiv:2407.21783.

\bibitem[{Gupta et~al.(2024)Gupta, Jegelka, Lopez{-}Paz, and Ahuja}]{GuptaJLA24}
Sharut Gupta, Stefanie Jegelka, David Lopez{-}Paz, and Kartik Ahuja. 2024.
\newblock \href {https://openreview.net/forum?id=8VPWfqtQMX} {Context is environment}.
\newblock In \emph{The Twelfth International Conference on Learning Representations, {ICLR} 2024, Vienna, Austria, May 7-11, 2024}. OpenReview.net.

\bibitem[{Hessel et~al.(2021)Hessel, Holtzman, Forbes, Le~Bras, and Choi}]{hessel-etal-2021-clipscore}
Jack Hessel, Ari Holtzman, Maxwell Forbes, Ronan Le~Bras, and Yejin Choi. 2021.
\newblock \href {https://doi.org/10.18653/v1/2021.emnlp-main.595} {{CLIPS}core: A reference-free evaluation metric for image captioning}.
\newblock In \emph{Proceedings of the 2021 Conference on Empirical Methods in Natural Language Processing}, pages 7514--7528, Online and Punta Cana, Dominican Republic. Association for Computational Linguistics.

\bibitem[{Hu et~al.(2024)Hu, Xu, Zhang, Ye, Yan, Zhang, Jin, Huang, and Zhou}]{hu2024mplugdocowl2highresolutioncompressingocrfree}
Anwen Hu, Haiyang Xu, Liang Zhang, Jiabo Ye, Ming Yan, Ji~Zhang, Qin Jin, Fei Huang, and Jingren Zhou. 2024.
\newblock \href {https://arxiv.org/abs/2409.03420} {mplug-docowl2: High-resolution compressing for ocr-free multi-page document understanding}.
\newblock \emph{Preprint}, arXiv:2409.03420.

\bibitem[{Hu et~al.(2022)Hu, Shen, Wallis, Allen{-}Zhu, Li, Wang, Wang, and Chen}]{LoRA}
Edward~J. Hu, Yelong Shen, Phillip Wallis, Zeyuan Allen{-}Zhu, Yuanzhi Li, Shean Wang, Lu~Wang, and Weizhu Chen. 2022.
\newblock \href {https://openreview.net/forum?id=nZeVKeeFYf9} {Lora: Low-rank adaptation of large language models}.
\newblock In \emph{The Tenth International Conference on Learning Representations, {ICLR} 2022, Virtual Event, April 25-29, 2022}. OpenReview.net.

\bibitem[{Ju et~al.(2021)Ju, Liu, Koh, Jin, Du, and Pan}]{DBLP:conf/emnlp/JuLKJDP21}
Jiaxin Ju, Ming Liu, Huan~Yee Koh, Yuan Jin, Lan Du, and Shirui Pan. 2021.
\newblock \href {https://doi.org/10.18653/v1/2021.findings-emnlp.345} {Leveraging information bottleneck for scientific document summarization}.
\newblock In \emph{Findings of the Association for Computational Linguistics: EMNLP 2021}, pages 4091--4098, Punta Cana, Dominican Republic. Association for Computational Linguistics.

\bibitem[{Kirillov et~al.(2023)Kirillov, Mintun, Ravi, Mao, Rolland, Gustafson, Xiao, Whitehead, Berg, Lo, Doll{\'{a}}r, and Girshick}]{DBLP:conf/iccv/KirillovMRMRGXW23}
Alexander Kirillov, Eric Mintun, Nikhila Ravi, Hanzi Mao, Chlo{\'{e}} Rolland, Laura Gustafson, Tete Xiao, Spencer Whitehead, Alexander~C. Berg, Wan{-}Yen Lo, Piotr Doll{\'{a}}r, and Ross~B. Girshick. 2023.
\newblock \href {https://openaccess.thecvf.com/content/ICCV2023/html/Kirillov_Segment_Anything_ICCV_2023_paper.html} {Segment anything}.
\newblock In \emph{{ICCV}}, pages 3992--4003. {IEEE}.

\bibitem[{Koh et~al.(2023)Koh, Salakhutdinov, and Fried}]{Koh_Grounding}
Jing~Yu Koh, Ruslan Salakhutdinov, and Daniel Fried. 2023.
\newblock \href {https://proceedings.mlr.press/v202/koh23a.html} {Grounding language models to images for multimodal inputs and outputs}.
\newblock In \emph{Proceedings of the 40th International Conference on Machine Learning}, volume 202 of \emph{Proceedings of Machine Learning Research}, pages 17283--17300. PMLR.

\bibitem[{Kuang et~al.(2021)Kuang, Sun, Li, Yue, Lin, Chen, Wei, Zhu, Gao, Zhang, Chen, Zhang, and Lin}]{MMOCR}
Zhanghui Kuang, Hongbin Sun, Zhizhong Li, Xiaoyu Yue, Tsui~Hin Lin, Jianyong Chen, Huaqiang Wei, Yiqin Zhu, Tong Gao, Wenwei Zhang, Kai Chen, Wayne Zhang, and Dahua Lin. 2021.
\newblock \href {https://doi.org/10.1145/3474085.3478328} {Mmocr: A comprehensive toolbox for text detection, recognition and understanding}.
\newblock In \emph{Proceedings of the 29th ACM International Conference on Multimedia}, MM '21, page 3791–3794, New York, NY, USA. Association for Computing Machinery.

\bibitem[{Laban et~al.(2022)Laban, Schnabel, Bennett, and Hearst}]{laban-etal-2022-summac}
Philippe Laban, Tobias Schnabel, Paul~N. Bennett, and Marti~A. Hearst. 2022.
\newblock \href {https://doi.org/10.1162/tacl_a_00453} {{S}umma{C}: Re-visiting {NLI}-based models for inconsistency detection in summarization}.
\newblock \emph{Transactions of the Association for Computational Linguistics}, 10:163--177.

\bibitem[{Lev et~al.(2019)Lev, Shmueli-Scheuer, Herzig, Jerbi, and Konopnicki}]{DBLP:conf/acl/LevSHJK19}
Guy Lev, Michal Shmueli-Scheuer, Jonathan Herzig, Achiya Jerbi, and David Konopnicki. 2019.
\newblock \href {https://doi.org/10.18653/v1/P19-1204} {{T}alk{S}umm: A dataset and scalable annotation method for scientific paper summarization based on conference talks}.
\newblock In \emph{Proceedings of the 57th Annual Meeting of the Association for Computational Linguistics}, pages 2125--2131, Florence, Italy. Association for Computational Linguistics.

\bibitem[{Li et~al.(2024)Li, Wang, Xu, Wang, Feng, Kong, and Liu}]{li-etal-2024-multimodal-arxiv}
Lei Li, Yuqi Wang, Runxin Xu, Peiyi Wang, Xiachong Feng, Lingpeng Kong, and Qi~Liu. 2024.
\newblock \href {https://doi.org/10.18653/v1/2024.acl-long.775} {Multimodal {A}r{X}iv: A dataset for improving scientific comprehension of large vision-language models}.
\newblock In \emph{Proceedings of the 62nd Annual Meeting of the Association for Computational Linguistics (Volume 1: Long Papers)}, pages 14369--14387, Bangkok, Thailand. Association for Computational Linguistics.

\bibitem[{Lin(2004)}]{lin-2004-rouge}
Chin-Yew Lin. 2004.
\newblock \href {https://aclanthology.org/W04-1013} {{ROUGE}: A package for automatic evaluation of summaries}.
\newblock In \emph{Text Summarization Branches Out}, pages 74--81, Barcelona, Spain. Association for Computational Linguistics.

\bibitem[{Liu et~al.(2024{\natexlab{a}})Liu, Wang, Loy, and Demberg}]{DBLP:conf/coling/PuWLD24}
Dongqi Liu, Yifan Wang, Jia Loy, and Vera Demberg. 2024{\natexlab{a}}.
\newblock \href {https://aclanthology.org/2024.lrec-main.1258/} {{S}ci{N}ews: From scholarly complexities to public narratives {--} a dataset for scientific news report generation}.
\newblock In \emph{Proceedings of the 2024 Joint International Conference on Computational Linguistics, Language Resources and Evaluation (LREC-COLING 2024)}, pages 14429--14444, Torino, Italia. ELRA and ICCL.

\bibitem[{Liu et~al.(2024{\natexlab{b}})Liu, Li, Li, and Lee}]{liu2023improvedllava}
Haotian Liu, Chunyuan Li, Yuheng Li, and Yong~Jae Lee. 2024{\natexlab{b}}.
\newblock \href {https://doi.org/10.1109/CVPR52733.2024.02484} {Improved baselines with visual instruction tuning}.
\newblock In \emph{{IEEE/CVF} Conference on Computer Vision and Pattern Recognition, {CVPR} 2024, Seattle, WA, USA, June 16-22, 2024}, pages 26286--26296. {IEEE}.

\bibitem[{Liu et~al.(2024{\natexlab{c}})Liu, Li, Li, Li, Zhang, Shen, and Lee}]{liu2024llavanext}
Haotian Liu, Chunyuan Li, Yuheng Li, Bo~Li, Yuanhan Zhang, Sheng Shen, and Yong~Jae Lee. 2024{\natexlab{c}}.
\newblock \href {https://llava-vl.github.io/blog/2024-01-30-llava-next/} {Llava-next: Improved reasoning, ocr, and world knowledge}.

\bibitem[{Liu et~al.(2023)Liu, Li, Wu, and Lee}]{DBLP:conf/nips/LiuLWL23a}
Haotian Liu, Chunyuan Li, Qingyang Wu, and Yong~Jae Lee. 2023.
\newblock \href {https://proceedings.neurips.cc/paper_files/paper/2023/file/6dcf277ea32ce3288914faf369fe6de0-Paper-Conference.pdf} {Visual instruction tuning}.
\newblock In \emph{Advances in Neural Information Processing Systems}, volume~36, pages 34892--34916. Curran Associates, Inc.

\bibitem[{Liu et~al.(2024{\natexlab{d}})Liu, Liu, Yu, Zhang, Jiang, Li, and Huang}]{DBLP:conf/acl/LiuLYZJLH24}
Ran Liu, Ming Liu, Min Yu, He~Zhang, Jianguo Jiang, Gang Li, and Weiqing Huang. 2024{\natexlab{d}}.
\newblock \href {https://doi.org/10.18653/v1/2024.findings-acl.574} {{S}um{S}urvey: An abstractive dataset of scientific survey papers for long document summarization}.
\newblock In \emph{Findings of the Association for Computational Linguistics: ACL 2024}, pages 9632--9651, Bangkok, Thailand. Association for Computational Linguistics.

\bibitem[{Liu et~al.(2024{\natexlab{e}})Liu, Duan, Zhang, Li, Zhang, Zhao, Yuan, Wang, He, Liu, Chen, and Lin}]{MMBench}
Yuan Liu, Haodong Duan, Yuanhan Zhang, Bo~Li, Songyang Zhang, Wangbo Zhao, Yike Yuan, Jiaqi Wang, Conghui He, Ziwei Liu, Kai Chen, and Dahua Lin. 2024{\natexlab{e}}.
\newblock \href {https://doi.org/10.1007/978-3-031-72658-3_13} {Mmbench: Is your multi-modal model an all-around player?}
\newblock In \emph{Computer Vision – ECCV 2024: 18th European Conference, Milan, Italy, September 29–October 4, 2024, Proceedings, Part VI}, page 216–233, Berlin, Heidelberg. Springer-Verlag.

\bibitem[{Lu et~al.(2024)Lu, Bansal, Xia, Liu, Li, Hajishirzi, Cheng, Chang, Galley, and Gao}]{lu2024mathvista}
Pan Lu, Hritik Bansal, Tony Xia, Jiacheng Liu, Chunyuan Li, Hannaneh Hajishirzi, Hao Cheng, Kai-Wei Chang, Michel Galley, and Jianfeng Gao. 2024.
\newblock \href {https://openreview.net/forum?id=KUNzEQMWU7} {Mathvista: Evaluating mathematical reasoning of foundation models in visual contexts}.
\newblock In \emph{The Twelfth International Conference on Learning Representations}.

\bibitem[{Luo et~al.(2024)Luo, Shen, Zhu, Zheng, Yu, and Yao}]{DBLP:conf/cvpr/LuoSZZYY24}
Chuwei Luo, Yufan Shen, Zhaoqing Zhu, Qi~Zheng, Zhi Yu, and Cong Yao. 2024.
\newblock \href {https://openaccess.thecvf.com/content/CVPR2024/html/Luo_LayoutLLM_Layout_Instruction_Tuning_with_Large_Language_Models_for_Document_CVPR_2024_paper.html} {Layoutllm: Layout instruction tuning with large language models for document understanding}.
\newblock In \emph{{CVPR}}, pages 15630--15640. {IEEE}.

\bibitem[{Masry et~al.(2022)Masry, Do, Tan, Joty, and Hoque}]{masry-etal-2022-chartqa}
Ahmed Masry, Xuan~Long Do, Jia~Qing Tan, Shafiq Joty, and Enamul Hoque. 2022.
\newblock \href {https://doi.org/10.18653/v1/2022.findings-acl.177} {{C}hart{QA}: A benchmark for question answering about charts with visual and logical reasoning}.
\newblock In \emph{Findings of the Association for Computational Linguistics: ACL 2022}, pages 2263--2279, Dublin, Ireland. Association for Computational Linguistics.

\bibitem[{Meta(2024)}]{llamaV}
AI~Meta. 2024.
\newblock \href {https://ai.meta.com/blog/llama-3-2-connect-2024-vision-edge-mobile-devices/} {Llama 3.2: Revolutionizing edge ai and vision with open, customizable models}.
\newblock \emph{Meta AI Blog. Retrieved December}, 20:2024.

\bibitem[{Min et~al.(2023)Min, Krishna, Lyu, Lewis, Yih, Koh, Iyyer, Zettlemoyer, and Hajishirzi}]{min-etal-2023-factscore}
Sewon Min, Kalpesh Krishna, Xinxi Lyu, Mike Lewis, Wen-tau Yih, Pang Koh, Mohit Iyyer, Luke Zettlemoyer, and Hannaneh Hajishirzi. 2023.
\newblock \href {https://doi.org/10.18653/v1/2023.emnlp-main.741} {{FA}ct{S}core: Fine-grained atomic evaluation of factual precision in long form text generation}.
\newblock In \emph{Proceedings of the 2023 Conference on Empirical Methods in Natural Language Processing}, pages 12076--12100, Singapore. Association for Computational Linguistics.

\bibitem[{OpenAI et~al.(2024)OpenAI, Achiam, Adler, Agarwal, and et~al.}]{openai2024gpt4technicalreport}
OpenAI, Josh Achiam, Steven Adler, Sandhini Agarwal, and Lama~Ahmad et~al. 2024.
\newblock \href {https://arxiv.org/abs/2303.08774} {Gpt-4 technical report}.
\newblock \emph{Preprint}, arXiv:2303.08774.

\bibitem[{Peng et~al.(2022)Peng, Pan, Wang, Luo, Zhang, Huang, Cao, Yin, Chen, Zhang, Feng, Sun, Tian, Wu, and Wang}]{ERNIE}
Qiming Peng, Yinxu Pan, Wenjin Wang, Bin Luo, Zhenyu Zhang, Zhengjie Huang, Yuhui Cao, Weichong Yin, Yongfeng Chen, Yin Zhang, Shikun Feng, Yu~Sun, Hao Tian, Hua Wu, and Haifeng Wang. 2022.
\newblock \href {https://doi.org/10.18653/V1/2022.FINDINGS-EMNLP.274} {Ernie-layout: Layout knowledge enhanced pre-training for visually-rich document understanding}.
\newblock In \emph{Findings of the Association for Computational Linguistics: {EMNLP} 2022, Abu Dhabi, United Arab Emirates, December 7-11, 2022}, pages 3744--3756. Association for Computational Linguistics.

\bibitem[{Post(2018)}]{post-2018-call}
Matt Post. 2018.
\newblock \href {https://www.aclweb.org/anthology/W18-6319} {A call for clarity in reporting {BLEU} scores}.
\newblock In \emph{Proceedings of the Third Conference on Machine Translation: Research Papers}, pages 186--191, Belgium, Brussels. Association for Computational Linguistics.

\bibitem[{Pramanick et~al.(2024)Pramanick, Chellappa, and Venugopalan}]{pramanick2024spiqa}
Shraman Pramanick, Rama Chellappa, and Subhashini Venugopalan. 2024.
\newblock \href {https://openreview.net/forum?id=h3lddsY5nf} {{SPIQA}: A dataset for multimodal question answering on scientific papers}.
\newblock In \emph{The Thirty-eight Conference on Neural Information Processing Systems Datasets and Benchmarks Track}.

\bibitem[{Rakitin et~al.(2024)Rakitin, Shchekotov, and Vetrov}]{rakitin2024regularized}
Denis Rakitin, Ivan Shchekotov, and Dmitry Vetrov. 2024.
\newblock \href {https://openreview.net/forum?id=Vg0wSHRnrn} {Regularized distribution matching distillation for one-step unpaired image-to-image translation}.
\newblock In \emph{ICML 2024 Workshop on Structured Probabilistic Inference {\&} Generative Modeling}.

\bibitem[{Saxena and Keller(2024)}]{saxena-keller-2024-select}
Rohit Saxena and Frank Keller. 2024.
\newblock \href {https://doi.org/10.18653/v1/2024.findings-naacl.218} {Select and summarize: Scene saliency for movie script summarization}.
\newblock In \emph{Findings of the Association for Computational Linguistics: NAACL 2024}, pages 3439--3455, Mexico City, Mexico. Association for Computational Linguistics.

\bibitem[{Sotudeh and Goharian(2022)}]{DBLP:conf/naacl/SotudehG22}
Sajad Sotudeh and Nazli Goharian. 2022.
\newblock \href {https://doi.org/10.18653/v1/2022.naacl-main.25} {{TSTR}: Too short to represent, summarize with details! intro-guided extended summary generation}.
\newblock In \emph{Proceedings of the 2022 Conference of the North American Chapter of the Association for Computational Linguistics: Human Language Technologies}, pages 325--335, Seattle, United States. Association for Computational Linguistics.

\bibitem[{Takeshita et~al.(2024)Takeshita, Green, Reinig, Eckert, and Ponzetto}]{DBLP:conf/naacl/TakeshitaGR0P24}
Sotaro Takeshita, Tommaso Green, Ines Reinig, Kai Eckert, and Simone Ponzetto. 2024.
\newblock \href {https://doi.org/10.18653/v1/2024.naacl-long.371} {{ACLS}um: A new dataset for aspect-based summarization of scientific publications}.
\newblock In \emph{Proceedings of the 2024 Conference of the North American Chapter of the Association for Computational Linguistics: Human Language Technologies (Volume 1: Long Papers)}, pages 6660--6675, Mexico City, Mexico. Association for Computational Linguistics.

\bibitem[{Tanaka et~al.(2023)Tanaka, Nishida, Nishida, Hasegawa, Saito, and Saito}]{DBLP:conf/aaai/TanakaNNHSS23}
Ryota Tanaka, Kyosuke Nishida, Kosuke Nishida, Taku Hasegawa, Itsumi Saito, and Kuniko Saito. 2023.
\newblock \href {https://doi.org/10.1609/aaai.v37i11.26598} {Slidevqa: A dataset for document visual question answering on multiple images}.
\newblock \emph{Proceedings of the AAAI Conference on Artificial Intelligence}, 37(11):13636--13645.

\bibitem[{Wang et~al.(2024{\natexlab{a}})Wang, Raman, Sibue, Ma, Babkin, Kaur, Pei, Nourbakhsh, and Liu}]{DBLP:conf/acl/0005RSMBKPNL24}
Dongsheng Wang, Natraj Raman, Mathieu Sibue, Zhiqiang Ma, Petr Babkin, Simerjot Kaur, Yulong Pei, Armineh Nourbakhsh, and Xiaomo Liu. 2024{\natexlab{a}}.
\newblock \href {https://doi.org/10.18653/v1/2024.acl-long.463} {{D}oc{LLM}: A layout-aware generative language model for multimodal document understanding}.
\newblock In \emph{Proceedings of the 62nd Annual Meeting of the Association for Computational Linguistics (Volume 1: Long Papers)}, pages 8529--8548, Bangkok, Thailand. Association for Computational Linguistics.

\bibitem[{Wang et~al.(2024{\natexlab{b}})Wang, Lv, Yu, Hong, Qi, Wang, Ji, Yang, Zhao, XiXuan, Xu, Chen, Xu, Li, Dong, Ding, and Tang}]{DBLP:journals/corr/abs-2311-03079}
Weihan Wang, Qingsong Lv, Wenmeng Yu, Wenyi Hong, Ji~Qi, Yan Wang, Junhui Ji, Zhuoyi Yang, Lei Zhao, Song XiXuan, Jiazheng Xu, Keqin Chen, Bin Xu, Juanzi Li, Yuxiao Dong, Ming Ding, and Jie Tang. 2024{\natexlab{b}}.
\newblock \href {https://proceedings.neurips.cc/paper_files/paper/2024/file/dc06d4d2792265fb5454a6092bfd5c6a-Paper-Conference.pdf} {Cogvlm: Visual expert for pretrained language models}.
\newblock In \emph{Advances in Neural Information Processing Systems}, volume~37, pages 121475--121499. Curran Associates, Inc.

\bibitem[{Yang et~al.(2024)Yang, Yang, Hui, Zheng, Yu, and et~al.}]{qwen2technicalreport}
An~Yang, Baosong Yang, Binyuan Hui, Bo~Zheng, Bowen Yu, and Chang~Zhou et~al. 2024.
\newblock \href {https://arxiv.org/abs/2407.10671} {Qwen2 technical report}.
\newblock \emph{Preprint}, arXiv:2407.10671.

\bibitem[{Yao et~al.(2024)Yao, Yu, Zhang, Wang, Cui, Zhu, Cai, Li, Zhao, He, Chen, Zhou, Zou, Zhang, Hu, Zheng, Zhou, Cai, Han, Zeng, Li, Liu, and Sun}]{yao2024minicpmvgpt4vlevelmllm}
Yuan Yao, Tianyu Yu, Ao~Zhang, Chongyi Wang, Junbo Cui, Hongji Zhu, Tianchi Cai, Haoyu Li, Weilin Zhao, Zhihui He, Qianyu Chen, Huarong Zhou, Zhensheng Zou, Haoye Zhang, Shengding Hu, Zhi Zheng, Jie Zhou, Jie Cai, Xu~Han, and 4 others. 2024.
\newblock \href {https://arxiv.org/abs/2408.01800} {Minicpm-v: A gpt-4v level mllm on your phone}.
\newblock \emph{Preprint}, arXiv:2408.01800.

\bibitem[{Yasunaga et~al.(2019)Yasunaga, Kasai, Zhang, Fabbri, Li, Friedman, and Radev}]{ScisummNet}
Michihiro Yasunaga, Jungo Kasai, Rui Zhang, Alexander~R. Fabbri, Irene Li, Dan Friedman, and Dragomir~R. Radev. 2019.
\newblock \href {https://doi.org/10.1609/AAAI.V33I01.33017386} {Scisummnet: {A} large annotated corpus and content-impact models for scientific paper summarization with citation networks}.
\newblock In \emph{The Thirty-Third {AAAI} Conference on Artificial Intelligence, {AAAI} 2019, The Thirty-First Innovative Applications of Artificial Intelligence Conference, {IAAI} 2019, The Ninth {AAAI} Symposium on Educational Advances in Artificial Intelligence, {EAAI} 2019, Honolulu, Hawaii, USA, January 27 - February 1, 2019}, pages 7386--7393. {AAAI} Press.

\bibitem[{Yu et~al.(2024)Yu, Sun, Zhang, Cui, Zhang, Cao, Wang, and Liu}]{Yu_2024_CVPR_CapsFusion}
Qiying Yu, Quan Sun, Xiaosong Zhang, Yufeng Cui, Fan Zhang, Yue Cao, Xinlong Wang, and Jingjing Liu. 2024.
\newblock \href {https://openaccess.thecvf.com/content/CVPR2024/html/Yu_CapsFusion_Rethinking_Image-Text_Data_at_Scale_CVPR_2024_paper.html} {Capsfusion: Rethinking image-text data at scale}.
\newblock In \emph{Proceedings of the IEEE/CVF Conference on Computer Vision and Pattern Recognition (CVPR)}, pages 14022--14032.

\bibitem[{Yue et~al.(2024)Yue, Ni, Zhang, Zheng, Liu, Zhang, Stevens, Jiang, Ren, Sun, Wei, Yu, Yuan, Sun, Yin, Zheng, Yang, Liu, Huang, Sun, Su, and Chen}]{Yue_2024_CVPR}
Xiang Yue, Yuansheng Ni, Kai Zhang, Tianyu Zheng, Ruoqi Liu, Ge~Zhang, Samuel Stevens, Dongfu Jiang, Weiming Ren, Yuxuan Sun, Cong Wei, Botao Yu, Ruibin Yuan, Renliang Sun, Ming Yin, Boyuan Zheng, Zhenzhu Yang, Yibo Liu, Wenhao Huang, and 3 others. 2024.
\newblock \href {https://openaccess.thecvf.com/content/CVPR2024/html/Yue_MMMU_A_Massive_Multi-discipline_Multimodal_Understanding_and_Reasoning_Benchmark_for_CVPR_2024_paper.html} {Mmmu: A massive multi-discipline multimodal understanding and reasoning benchmark for expert agi}.
\newblock In \emph{Proceedings of the IEEE/CVF Conference on Computer Vision and Pattern Recognition (CVPR)}, pages 9556--9567.

\bibitem[{Zhang et~al.(2020)Zhang, Kishore, Wu, Weinberger, and Artzi}]{bert-score}
Tianyi Zhang, Varsha Kishore, Felix Wu, Kilian~Q. Weinberger, and Yoav Artzi. 2020.
\newblock \href {https://openreview.net/forum?id=SkeHuCVFDr} {Bertscore: Evaluating text generation with bert}.
\newblock In \emph{International Conference on Learning Representations}.

\bibitem[{Zheng et~al.(2024)Zheng, Feng, Si, She, Lin, Jiang, and Wang}]{zheng-etal-2024-multimodal}
Mingyu Zheng, Xinwei Feng, Qingyi Si, Qiaoqiao She, Zheng Lin, Wenbin Jiang, and Weiping Wang. 2024.
\newblock \href {https://doi.org/10.18653/v1/2024.acl-long.493} {Multimodal table understanding}.
\newblock In \emph{Proceedings of the 62nd Annual Meeting of the Association for Computational Linguistics (Volume 1: Long Papers)}, pages 9102--9124, Bangkok, Thailand. Association for Computational Linguistics.

\end{thebibliography}

\clearpage

\appendix
\begin{table}[t]
\centering
\scalebox{1}{
\begin{tabular}{lcccc}
\toprule
\textbf{Methods} & \textbf{SummaC} & \textbf{FactScore}\\
\midrule
Llama-3.2 FT & 21.55 & 76.67  \\
MiniCPM CoT & 23.03 & 92.10\\
GPT-4o & 23.25 & 94.44 \\ 
Ours &  25.41 & 94.67 \\ 
\bottomrule
\end{tabular}
}
\caption{Results of automatic evaluation of factual
consistency on the best model in each category}
\label{tab:factuality_result}
\end{table}
\section{Factuality Evaluation} \label{sec:factuality}
To evaluate the performance of our method in generating factually correct summaries, we compute two text-based metrics: SummaC Conv~\citep{laban-etal-2022-summac} and FActScore~\citep{min-etal-2023-factscore} on the best models in each category. Following common practice in long-document summarization evaluation~\citep{fabbri-etal-2022-qafacteval,saxena-keller-2024-select}, we treat the reference summary as the ground truth (instead of the original document, which is poster image) when computing these metrics. \cref{tab:factuality_result} presents the results for both metrics on the generated summaries.

Both metrics perform poorly, as they are not specialized for scientific text. SummaC scores were substantially low, while FActScore showed extremely high values, indicating failures in natural language inference and atomic fact extraction for scientific text. We found factuality evaluation to be a challenging task in this domain, highlighting the need for new methods to measure factual accuracy in multimodal scientific documents such as posters.

\section{Prompt Templates} 
\label{sec:templates}
\begin{figure}[htpb]
    \centering
    \begin{tcolorbox}[
        colback=white,       
        colframe=black,   
        title=Prompt Template for Zero-Shot,
        fonttitle=\bfseries,
        halign title=center,
        width=0.9\columnwidth,
        boxrule=1pt,          
        coltitle=black,       
        colbacktitle=white   
    ]
    Write an abstract for an AI conference paper for the given research poster image.
    \end{tcolorbox}
\end{figure}

\begin{figure}[htpb]
    \centering
    \scalebox{0.9}{
    \begin{tcolorbox}[
        colback=white,        
        colframe=black,       
        title=Prompt Template for CoT,
        fonttitle=\bfseries,  
        halign title=center,  
        width=1\columnwidth,  
        boxrule=1pt,          
        coltitle=black,       
        colbacktitle=white    
    ]
Analyze the research poster image step by step.\\
First, identify the title and main research problem.\\
Then, briefly describe the methodology used. \\
Next, summarize the key findings or results.\\
Finally, note the conclusion or implications.\\
Using this information, write an abstract for the given research poster image.
    \end{tcolorbox}
    }
\end{figure}
\begin{figure}[htpb]
    \centering
    \begin{tcolorbox}[
        colback=white,       
        colframe=black,      
        title=Prompt Template for Local Summary,
        fonttitle=\bfseries, 
        halign title=center, 
        width=1\columnwidth, 
        boxrule=1pt,         
        coltitle=black,      
        colbacktitle=white   
    ]
    Describe all the text, tables, figures, and equations in the image.
    \end{tcolorbox}
\end{figure}
\begin{figure}[t!]
  \centering
\includegraphics[width=0.9\columnwidth]{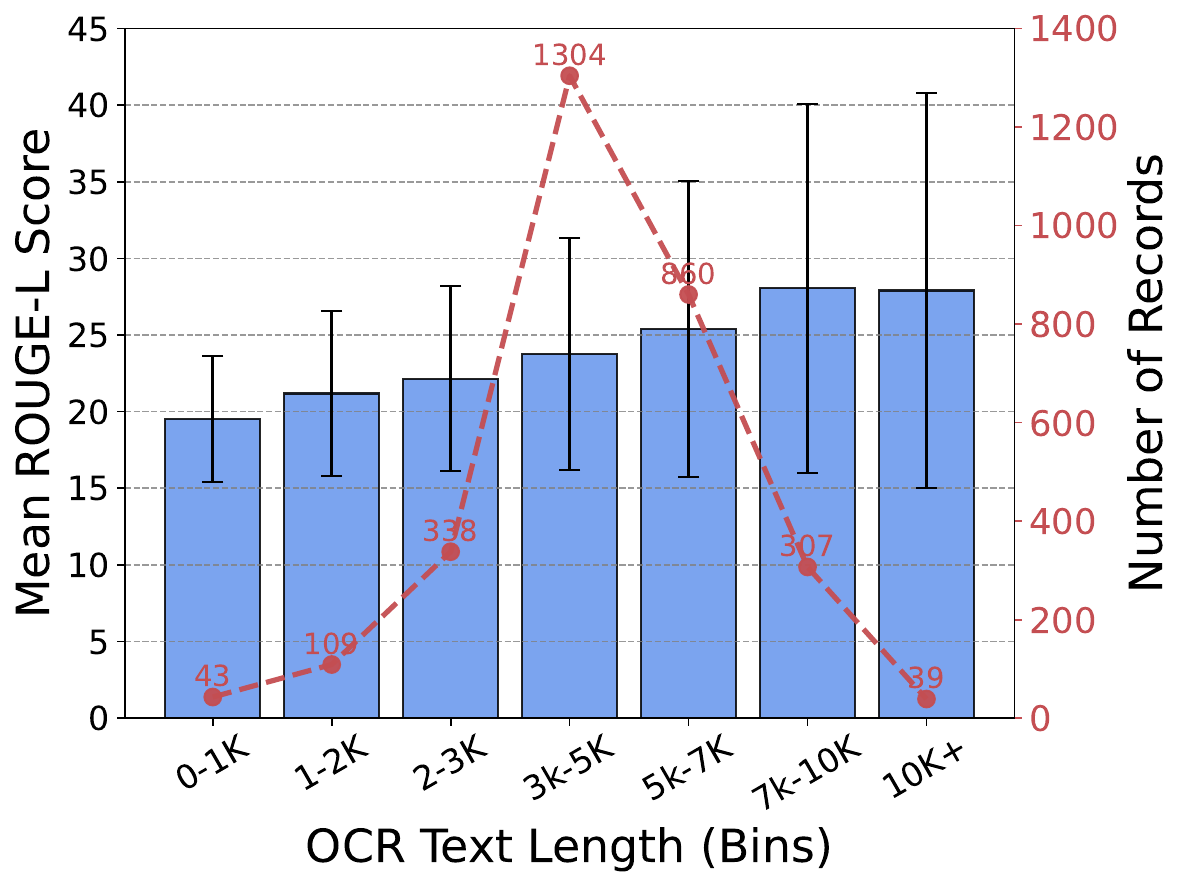}
\vspace{-2ex}
  \caption{Effect of text present in the poster on summarization. We report mean ROUGE-L scores for different OCR-extracted character-length bins. The red dashed line represents the number of posters in each bin.}
  \label{fig:ocr_rouge_l}
\end{figure}
\section{Effect of Poster Text Content on Summarization Performance}
\label{sec:text_rl}

To investigate whether posters with a high amount of text result in better summarization performance, we analyze the relationship between OCR-extracted text length and ROUGE-L scores using our \textsc{Segment \& Summarize} method. Specifically, we use MMOCR to extract text from each poster and compute its total length in characters (not in tokens). 

\cref{fig:ocr_rouge_l} presents the mean ROUGE-L scores across different OCR text-length bins. The dotted line represents the number of posters in each text-length bin. We observe that summarization performance tends to improve as the amount of text in poster increases. However, the correlation remains weak (\textit{Pearson} $r=0.213$, \textit{Spearman} $r=0.210$), suggesting that text in poster alone is not a strong predictor of summarization quality.
Low performance in posters with minimal text also highlights the need for more robust multimodal understanding of figures, charts, equations, and tables.
\begin{figure}[t!]
  \centering
\includegraphics[width=0.9\columnwidth]{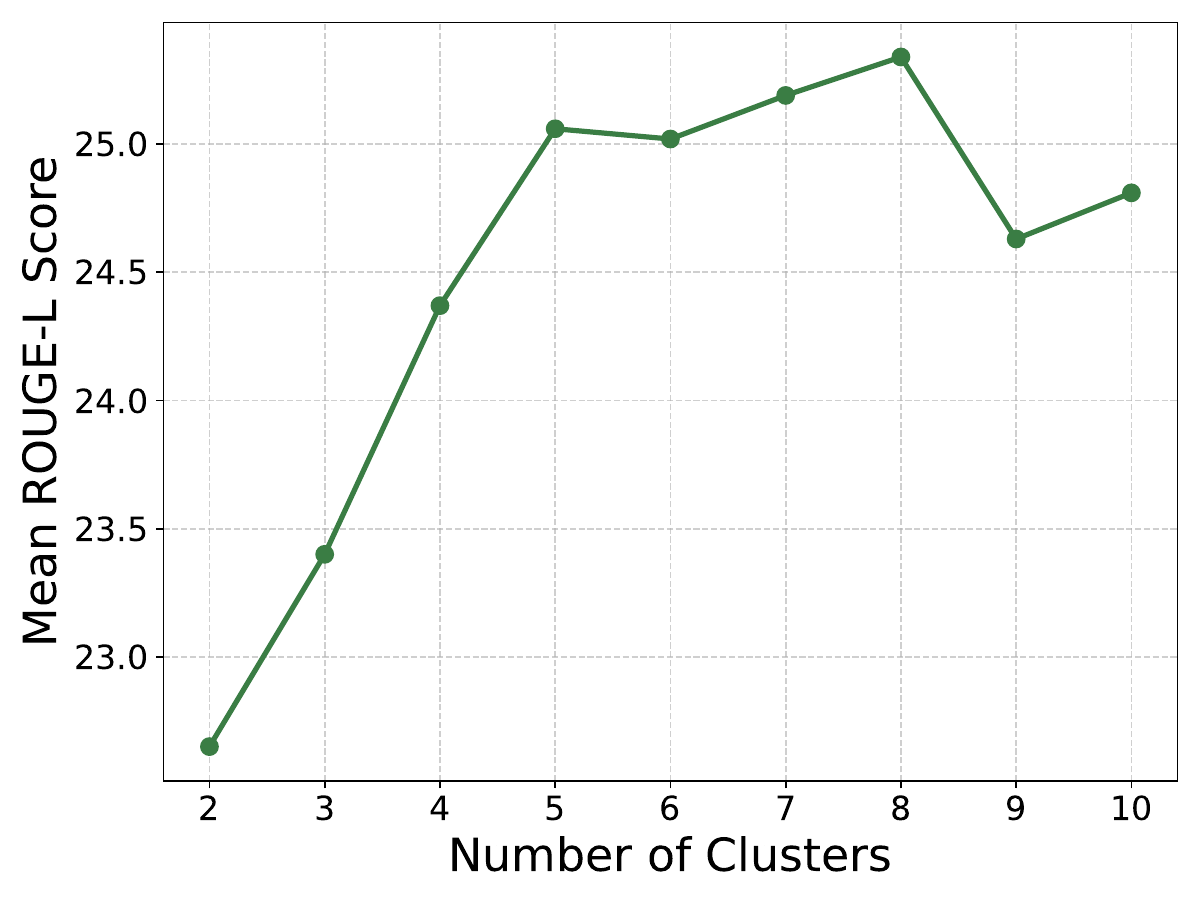}
\vspace{-2ex}
  \caption{Effect of varying the number of clusters on ROUGE-L performance on \method}
  \label{fig:cluser_rouge_plot}
\end{figure}
\section{Selecting the Number of Clusters}
\label{sec:number_of_cluster}
To select the number of clusters ($k$) for our \method, we conducted an empirical analysis on a subset of 100 posters from the validation set, varying the number of clusters from 2 to 10. \cref{fig:cluser_rouge_plot} presents the mean ROUGE-L score for each cluster configuration. In these experiments, the local and global summarization components remained fixed.

We observe that the best performance is achieved at $k=8$ which was used in our final experiments.  Additionally, we limit the maximum number of clusters to 10 in the analysis to keep the inference time of our local summarization manageable.
\begin{table}[t]
\centering
\scalebox{0.9}{
\begin{tabular}{l l}
\toprule
Model & Version\\
\midrule
GPT-4o & gpt-4o-2024-08-06 \\
Gemini 2.0 & gemini-2.0-flash-exp\\
Claude 3.5 Sonnet & claude-3-5-sonnet-20241022\\
\bottomrule
\end{tabular}
}
\caption{Details of the closed-sourced models.}
\label{tab:model_details}
\end{table}
\section{Additional Experiment Details}
\label{experiment_details_extra}
\cref{tab:model_details} summarizes the versions of the closed-source models used in our experiments. For fine-tuning, we use a learning rate of $1\times 10^{-4}$ with the Adam optimizer ($\beta_1 = 0.9, \beta_2 = 0.999, \epsilon = 1\times 10^{-8}$) and a cosine learning rate schedule. We employ LoRA with rank $r = 8$, $\alpha = 8$, and a dropout rate of 0.1.

All images are processed and scaled by the respective model’s image processor for model specific sizes. In the case of closed-source models, we scale each image to a maximum width of 2048 while preserving the original aspect ratio due to size limitations.  All the models were trained using 2 A100 GPU with 80GB memory. We used the Huggingface $evaluate$ library for the implementation of the metrics.

\section{Dataset Examples with Model Summaries}
\label{data_sample_summ}
\begin{table*}[t]
\centering
\scalebox{1}{\tabcolsep=3pt
\begin{threeparttable}
\begin{tabular}{|p{3cm}|p{10cm}|}
\hline
\multicolumn{2}{|c|}{
\includegraphics[width=0.9\textwidth]{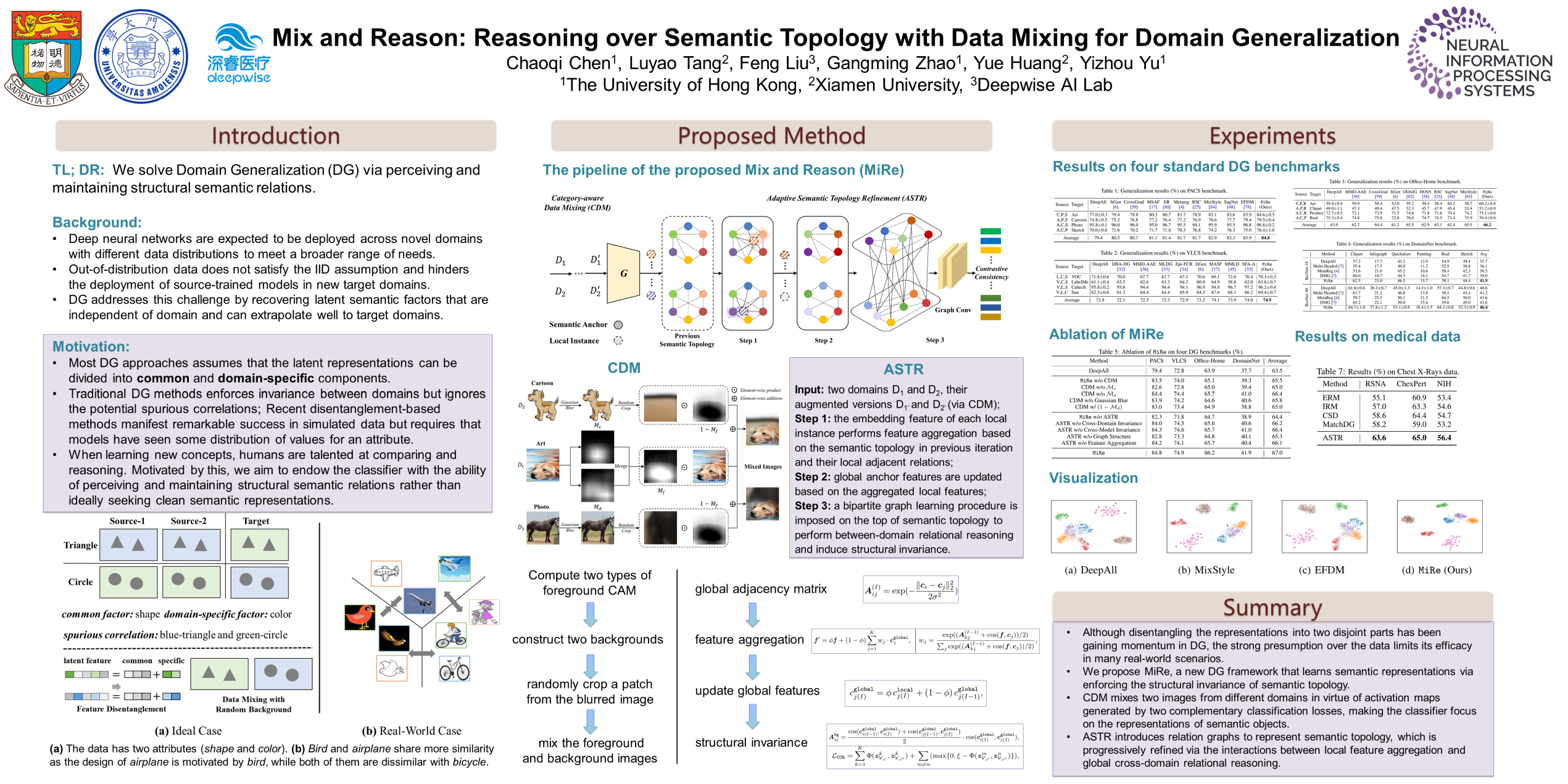}
}
\\
\hline
\scriptsize \textbf{Model} & \scriptsize \textbf{Output}\\
\hline
\scriptsize Reference & \scriptsize Domain generalization (DG) enables generalizing a learning machine from multiple seen source domains to an unseen target one. The general objective of DG methods is to learn semantic representations that are independent of domain labels, which is theoretically sound but empirically challenged due to the complex mixture of common and domain-specific factors. Although disentangling the representations into two disjoint parts has been gaining momentum in DG, the strong presumption over the data limits its efficacy in many real-world scenarios. In this paper, we propose Mix and Reason (MiRe), a new DG framework that learns semantic representations via enforcing the structural invariance of semantic topology. MiRe consists of two key components, namely,  Category-aware Data Mixing (CDM) and Adaptive Semantic Topology Refinement (ASTR). CDM mixes two images from different domains in virtue of activation maps generated by two complementary classification losses, making the classifier focus on the representations of semantic objects. ASTR introduces relation graphs to represent semantic topology, which is progressively refined via the interactions between local feature aggregation and global cross-domain relational reasoning. Experiments on multiple DG benchmarks validate the effectiveness and robustness of the proposed MiRe.  \\
\hline
\scriptsize MiniCPM CoT & \scriptsize  The paper is about Mix and Reason, a methodology for domain generalization that reasons over semantic topology to mix data from different domains and construct new target domains. It addresses the problem of limited training data by leveraging the IID assumption and the representations learned from the source domain to perform well in the target domain. The proposed methodology involves a pipeline of the proposed Mix and Reason (MiRe) component, which takes the adaptive semantic topology refinement (ASTR) component as its backbone. The key results show that the proposed methodology outperforms state-of-the-art methods on four benchmark datasets, including CelebA, CUB, Stanford 2007, and COCO.\\
\hline
\scriptsize  Llama-3.2-11B-V FT & \scriptsize  Deep neural networks (DNNs) are expected to be deployed across novel domains. However, out-of-distribution (OOD) data does not satisfy the assumption and hinders DNNs from generalizing well to target domains. In this paper, we propose Mix and Reason (MIRa) for domain generalization (DG) via perceiving and reasoning over semantic topology. Specifically, MIRa divides input common and domain-specific components and captures their potential spurious correlations. Reasoning over semantic topology enables MIRa to disentangle attribute and domain-specific representations and learn domain-invariant representations. Extensive experiments on four standard DG benchmarks show that MIRa achieves state-of-the-art performance. Besides, MIRa can be applied to many DG methods and significantly improve their performance.\\
\hline
\scriptsize  GPT-4o & \scriptsize Domain Generalization (DG) for deep neural networks poses a significant challenge in adapting models to varied out-of-distribution data. This paper introduces "Mix and Reason" (MiRe), a novel framework leveraging semantic topology reasoning and data mixing to enhance domain generalization. MiRe employs a two-part approach, utilizing Anchored Semantic Topology Reasoning (ASTR) to retain relational features across domains and Contrastive Data Mixing (CDM) to generate robust feature representations. Experiments were conducted on four standard DG benchmarks, demonstrating substantial improvements in model adaptability and performance. Additionally, results on medical datasets underscore the framework\'s efficacy in maintaining domain-invariant characteristics across diverse conditions. This research highlights the strength of integrating semantic relationships with innovative data augmentation techniques, offering a scalable solution for realistic multi-domain scenarios. \\
\hline
\scriptsize  \method & \scriptsize  Domain Generalization (DG) is a critical challenge in machine learning, where models are expected to generalize well across novel domains with different data distributions. Existing DG methods often rely on strong assumptions about the data or ignore potential spurious correlations. In this work, we propose a novel DG framework, Mix and Reason (MiRe), which learns semantic representations by enforcing the structural invariance of semantic topology. MiRe consists of two key components: Category-Aware Data Mixing (CDM) and Adaptive Logic Semantic Topology Refinement (ASTR). CDM mixes two images from different domains using activation maps generated by complementary classification losses, focusing the classifier on semantic object representations. ASTR introduces relation graphs to represent semantic topology, which is progressively refined via interactions between local feature aggregation and global cross-domain relational reasoning. We evaluate MiRe on four standard DG benchmarks and medical data, achieving state-of-the-art performance and outperforming existing methods. Our results demonstrate the effectiveness of MiRe in handling domain shifts and provide insights into how to better perceive and maintain structural semantic relations in machine learning.\\
\hline
\end{tabular}
\end{threeparttable}
}
\vspace*{-.2cm}
\caption{Sample of poster image from \citet{example1_paper} with gold reference and model generated summaries}
\label{tab:example1}
\end{table*}

\begin{table*}[t]
\centering
\scalebox{1}{\tabcolsep=3pt
\begin{threeparttable}
\begin{tabular}{|p{3cm}|p{11cm}|}
\hline
\multicolumn{2}{|c|}{
\includegraphics[width=0.8\textwidth]{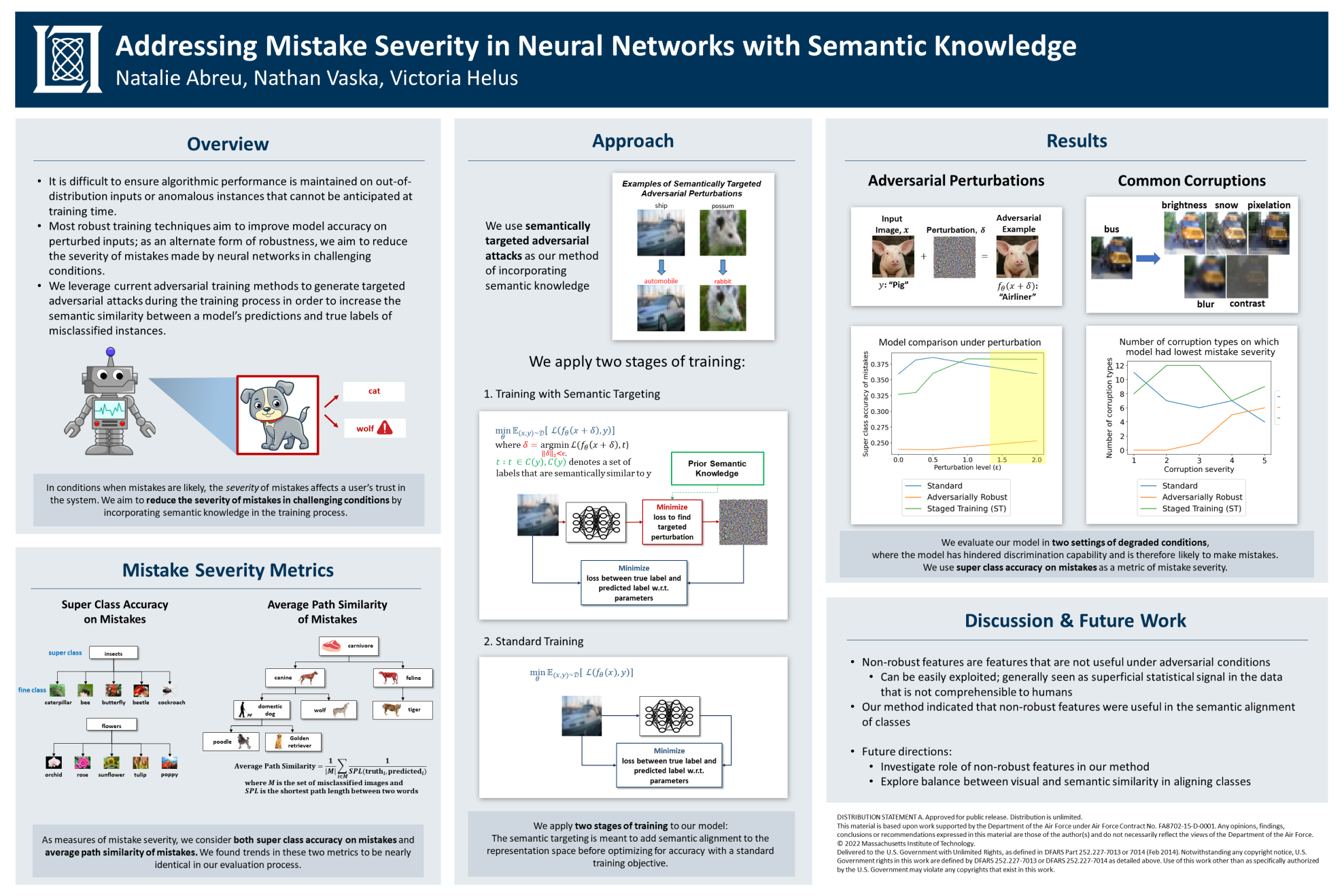}
}
\\
\hline
\scriptsize \textbf{Model} & \scriptsize \textbf{Output}\\
\hline
\scriptsize Reference & \scriptsize Robustness in deep neural networks and machine learning algorithms in general is an open research challenge. In particular, it is difficult to ensure algorithmic performance is maintained on out-of-distribution inputs or anomalous instances that cannot be anticipated at training time. Embodied agents will be deployed in these conditions, and are likely to make incorrect predictions. An agent will be viewed as untrustworthy unless it can maintain its performance in dynamic environments. Most robust training techniques aim to improve model accuracy on perturbed inputs; as an alternate form of robustness, we aim to reduce the severity of mistakes made by neural networks in challenging conditions. We leverage current adversarial training methods to generate targeted adversarial attacks during the training process in order to increase the semantic similarity between a model's predictions and true labels of misclassified instances. Results demonstrate that our approach performs better with respect to mistake severity compared to standard and adversarially trained models. We also find an intriguing role that non-robust features play with regards to semantic similarity. \\
\hline
\scriptsize MiniCPM CoT & \scriptsize  The paper is about the problem of mistake severity in neural networks and the use of semantic knowledge to address it. The research problem is the inability of neural networks to accurately predict the severity of mistakes, leading to poor model accuracy. The methodology involves incorporating semantic knowledge into the training process to generate targeted adversarial attacks. The key results show that using semantic knowledge to guide the attacks leads to better model accuracy and improved performance on benchmarks. The conclusion suggests that the use of semantic knowledge is a promising approach to addressing the problem of mistake severity in neural networks.\\
\hline
\scriptsize  Llama-3.2-11B-V FT & \scriptsize It is difficult to ensure algorithmic performance is maintained at out-of-distribution (OOD) training time for neural networks due to the presence of perturbed inputs. An alternate form of robustness, we aim to reduce the conditions in which mistakes are made by neural networks. We leverage semantic similarity between a model's predictions and labels of OOD data as a measure of mistake severity. We propose a two-stage training method that generates targeted perturbations of semantic similarity between a model's predictions and true labels of OOD data. Our experiments show that our method improves upon state-of-the-art methods in aligning class similarity between a model's predictions and true labels of OOD data. Our code is available at https://github.com/natale-abreu/semi-targeted-adversarial-training.git.\\
\hline
\scriptsize  GPT-4o & \scriptsize This paper addresses the challenge of reducing mistake severity in neural networks under adversarial and anomalous conditions by incorporating semantic knowledge into the training process. Traditional robust training techniques typically focus on improving model accuracy on perturbed inputs, but often overlook the semantic alignment between model predictions and true labels. We propose a novel method utilizing semantically targeted adversarial attacks to train neural networks, aiming to enhance semantic similarity and reduce the impact of severe errors.Our approach encompasses a two-stage training process: firstly integrating semantic knowledge through targeted adversarial perturbations, followed by standard training protocols. The methodology was tested using metrics such as super class accuracy on mistakes and average path similarity of mistakes, which measure semantic closeness between mistaken and true labels. This research highlights the value of non-robust features in capturing semantic alignments and suggests future exploration into balancing visual and semantic similarities for enhanced model reliability. \\
\hline
\scriptsize  \method & \scriptsize  Ensuring algorithmic performance on out-of-distribution inputs or anomalous instances is a significant challenge in neural networks. Traditional robust training techniques focus on improving model accuracy on perturbed inputs, whereas we propose an alternate form of robustness by reducing the severity of mistakes made by neural networks in challenging conditions. We leverage current adversarial training methods to generate targeted adversarial attacks during the training process, increasing the semantic similarity between a model's predictions and true labels of misclassified instances. Our approach involves two stages of training: Semantic Targeting and Prior Semantic Knowledge. We demonstrate that by incorporating semantic knowledge in the training process, we can reduce the severity of mistakes in challenging conditions, thereby improving user trust in the system. Our results show that the proposed method outperforms traditional robust training techniques in terms of reducing mistake severity, making it a promising approach for addressing mistake severity in neural networks.\\
\hline
\end{tabular}
\end{threeparttable}
}
\vspace*{-.2cm}
\caption{Sample of poster image from the work \citet{example2_paper} with gold reference and model generated summaries}
\label{tab:example2}
\end{table*}
\end{document}